\documentclass[journal]{IEEEtran}
\usepackage{cite}
\usepackage{amsmath,amssymb,amsfonts}
\usepackage{algorithmic}
\usepackage{graphicx}
\usepackage{graphics}
\usepackage{textcomp}
\usepackage{xcolor}
\usepackage{multirow}
\usepackage{placeins}
\usepackage{caption}
 \usepackage{float}
\usepackage{subfig}
\usepackage{adjustbox}
\usepackage{pifont}
\usepackage{epstopdf}    % Included EPS files automatically converted to PDF to include with pdflatex

\usepackage{makecell}
\setcellgapes{3pt}

\newcommand\tstrut{\rule{0pt}{2.4ex}}
\newcommand\bstrut{\rule[-1.0ex]{0pt}{0pt}}

\newcommand*\rot{\rotatebox{90}}
\def\BibTeX{{\rm B\kern-.05em{\sc i\kern-.025em b}\kern-.08em
    T\kern-.1667em\lower.7ex\hbox{E}\kern-.125emX}}
\begin{document}

\title{Efficient Context Integration through Factorized Pyramidal Learning for Ultra-Lightweight Semantic Segmentation}

%\author{\IEEEauthorblockN{Nadeem Atif \: \: \: \: Saquib Mazhar \: \: \: \: M. K. Bhuyan \: \: \: \: Shaik Rafi Ahamad} 

\author{Nadeem Atif, Saquib Mazhar, Debajit Sarma, M. K. Bhuyan and Shaik Rafi Ahamed

\IEEEauthorblockA{\textit{Dept. of Electronics and Electrical Engineerning, IIT Guwahati} \\
%\textit{IIT, Guwahati}\\
Guwahati- 781039, India \\
\{atif176102103, saquibmazhar, s.debajit, mkb, rafiahamed\}@iitg.ac.in}
%\IEEEauthorblockA{\IEEEauthorrefmark{*}balajihosur944@gmail.com}
}

\maketitle

\begin{abstract}
Semantic segmentation is a pixel-level prediction task to classify each pixel of the input image. Deep learning models, such as convolutional neural networks (CNNs), have been extremely successful in achieving excellent performances in this domain. However, mobile application, such as autonomous driving, demand real-time processing of incoming stream of images. Hence, achieving efficient architectures along with enhanced accuracy is of paramount importance. Since, accuracy and model size of CNNs are intrinsically contentious in nature, the challenge is to achieve a decent trade-off between accuracy and model size. To address this, we propose a novel Factorized Pyramidal Learning (FPL) module to aggregate rich contextual information in an efficient manner. On one hand, it uses a bank of convolutional filters with multiple dilation rates which leads to multi-scale context aggregation; crucial in achieving better accuracy. On the other hand, parameters are reduced by a careful factorization of the employed filters; crucial in achieving lightweight models. Moreover, we decompose the spatial pyramid into two stages which enables a simple and efficient feature fusion within the module to solve the notorious checkerboard effect. We also design a dedicated Feature-Image Reinforcement (FIR) unit to carry out the fusion operation of shallow and deep features with the downsampled versions of the input image. This gives an accuracy enhancement without increasing model parameters. Based on the FPL module and FIR unit, we propose an ultra-lightweight real-time network, called FPLNet, which achieves state-of-the-art accuracy-efficiency trade-off. More specifically, with only less than 0.5 million parameters, the proposed network achieves 66.93\% and 66.28\% mIoU on Cityscapes validation and test set, respectively. Moreover, FPLNet has a processing speed of 95.5 frames per second (FPS).
\end{abstract}


\section{Introduction}
\IEEEPARstart{S}{emantic} segmentation is a computer vision task that deals with classifying each and every single pixel of an image. Being a pixel-level prediction task, it is one of the most challenging tasks in visual recognition domain \cite{long2015fully, badrinarayanan2017segnet, chen2018encoder}. The rapid growth of deep learning (a sub-field of machine learning) during the last decade has revolutionized the field of computer vision \cite{szegedy2014going, simonyan2014very, he2016deep, krizhevsky2012imagenet, chollet2017xception}. Consequently, semantic segmentation has also greatly benefited from the recent developments in convolutional neural networks (CNNs) which is a deep learning model \cite{hong2021deep, yu2021bisenet, wu2020cgnet}. Autonomous driving, robotics, virtual and augmented reality, aerial imagery are some of the application areas of semantic segmentation. Out of these, autonomous driving is currently a hot topic of research both in industries and in academia \cite{cordts2016cityscapes, brostow2009semantic}. 
As a result, massive amounts of works have been published in literature reporting great performances in terms of accuracy of segmentation models \cite{atif2019review}. However, in addition to high accuracy, mobile application scenarios such as driverless cars and drones demand small size models for real-time processing of incoming stream of images. So, it is equally important to consider the model size while designing segmentation networks for resource-constrained devices. This prompted researchers to scale down the model sizes and a number of lightweight networks have been developed \cite{paszke2016enet, mehta2018espnet} as a result. However, these small model sizes were achieved at the cost of excessive reduction of inference accuracy; not suitable for practical application. It is therefore extremely important to keep both the attributes, i.e., accuracy and number of parameters in mind while designing semantic segmentation networks. In other words, achieving a decent trade-off between accuracy and model size is crucial in designing networks for resource-constrained real-time applications. As a result, it is currently a very promising area of research \cite{wu2020cgnet, nirkin2021hyperseg, li2019dabnet, poudel2018contextnet, lo2019efficient}.

\begin{figure*}
  \centering
  \includegraphics[width=0.9\linewidth]{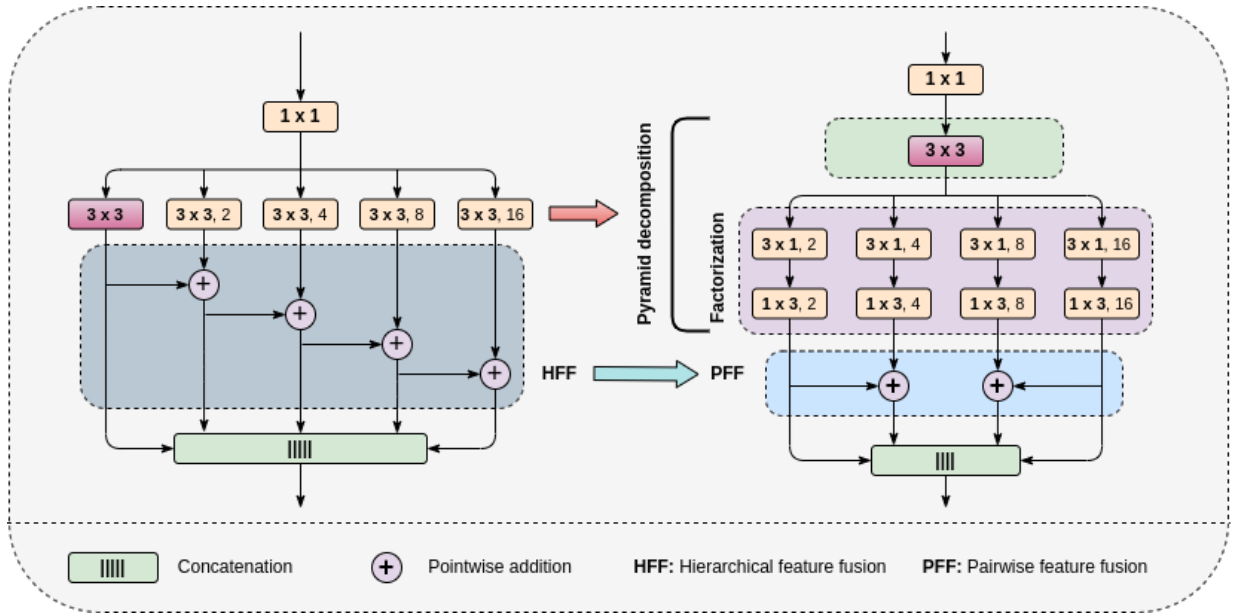}
  \caption{\small Comparison of skeletal structure of ESP and FPL module. The comma separated numbers in the convolutional blocks are the dilation factors.}
  \label{espvsfpl}
\end{figure*}

The overall accuracy of a network depends upon two types of information; high-level contextual information and low-level spatial information \cite{yu2018bisenet}. The high-level contextual information is crucial in producing globally consistent segmentation whereas low-level spatial information preserve the finer local details \cite{yu2021bisenet, kumaar2021cabinet}. Many works have recently been done to exploit these two types of information separately in two parallel branches \cite{yu2021bisenet, hong2021deep, kumaar2021cabinet}. Although, the resulting performances of these two branched approaches are quite exciting, the corresponding models still remain bulky having millions of parameters. So, achieving a decent accuracy-efficiency trade-off demands an efficient extraction of multi-scale context. To achieve this goal, we propose a Factorized Pyramidal Learning (FPL) module which is inspired by the basic block of the classical lightweight ESPNet \cite{mehta2018espnet}. ESP module employs a set of dilated convolutions with different dilation rates. To reduce parameters, the ESP module excessively compresses the channel dimension of feature blocks (1/5 of that of input block). This causes a significant information loss along the channel dimension. To address this, the spatial pyramid in FPL is decomposed into two stages. The first stage employs a single conventional convolution layer while the second stage is comprised of a bank of factorized filters. This decomposition allows us to feed 25\% more channels (1/4 of input channels) to the convolutional filters in comparison to ESP module. Although, the pyramid decomposition increases information flow along the channel dimension, it makes the module bulky compared to ESP module. So, to counter this effect, we employ factorization of convolutional filters in the second stage of spatial pyramid which results in huge parameter saving. We empirically found that the factorization of filters in the second stage has very little effect on the accuracy. Moreover, this two-step transformation enables a simple solution to the problem of gridding or checkerboard effect; caused by direct concatenation of outputs of a set dilated convolutions. More specifically, we propose an efficient Pairwise Feature Fusion (PFF) involving only two feature tensors as opposed to Hierarchical feature fusion (HFF) of ESPNet which involves all the feature tensors of the module. A comparison of the skeletal structure of ESP and FPL module is shown in Fig. \ref{espvsfpl}. 

To improve the information flow, feature maps from the first and the last block of an encoder stage \cite{lo2019efficient} are fused. To enhance the accuracy, input image insertion is also a common practice with only a slight increment in number of parameters \cite{mehta2018espnet, li2019dabnet}. The intra-stage feature fusion and the input image insertion is usually done by concatenation of feature tensors and downsampled (usually by pooling operation) version of input image. However, in this work, we design a dedicated Feature-Image Reinforcement (FIR) unit to carry out this fusion operation which offers a simple yet effective solution without any increase in network parameters. We also design a simple and sequential asymmetric decoder that helps in recovering the low-level details that are lost during the downsampling operations in encoder. The decoder employed in the proposed network is free from any encoder to decoder skip connections and it is kept purely sequential. This keeps the decoder fairly simple by eliminating the requirement to store the feature maps from different stages of encoder. This fares well in terms of Graphics Processing Unit (GPU) memory utilization and ultimately leads to better efficiency. 

Based on the FPL module, FIR unit and the asymmetric decoder, we propose an ultra-lightweight network, called FPLNet, to perform semantic segmentation in real time. 

The main contributions of this work are summarized as follows--
\begin{itemize}
    \item A novel FPL module is proposed which is inspired by the classical ESP module. On one hand, it harvests rich contextual information by capturing context at multiple scales and on the other hand, causes a huge parameter saving by introducing factorization into spatial pyramid of filters.
    \item The pyramidal decomposition enables a simpler solution to address the gridding artifact problem. The proposed solution is called Pairwise Feature Fusion (PFF) which required only two feature block at a time.
    \item A simple Feature-Image Reinforcement (FIR) unit has been proposed to jointly fuse deep and shallow features with the image at the end of each stage. It improves the accuracy further without requiring additional network parameters.
    \item Based on the FPL module, FIR unit and a simple asymmetric decoder, we propose an ultra-lightweight network, called FPLNet, which achieves state-of-the-art accuracy-efficiency trade-off with regards to lightweight networks.
\end{itemize}

The rest of the paper is organised as follows. In section \ref{related-work}, a brief survey of the related works is presented. Section \ref{related-work} also briefly presents the gap found in the existing literature which the proposed work fills. Section \ref{proposed_network} presents the basic building blocks of the proposed network which includes initial module, FPL module and the FIR unit. The overall architecture design of the corresponding FPLNet is also presented in detail in section \ref{proposed_network}. In section \ref{experiments}, various experiment related results and discussions are presented including dataset, training details, performance analysis and comparison of our FPLNet with other state-of-the-art networks. Section \ref{experiments}, also presents a detailed ablation study of various important design choices. Finally, section \ref{conclusion} concludes the paper.

\section{Related Work} \label{related-work}
To be consistent with our objective of developing small size model capable of achieving decent accuracy, we categorize the existing works based on their corresponding model sizes. More specifically, we divide the existing networks into four broad categories. Large-scale, mid-scale, lightweight and ultra-lightweight networks. Large-scale networks include models having equal to or more than 5 million parameters. Similarly, mid-scale, lightweight and ultra-lightweight networks include models having 1-5, 0.5-1 and less than 0.5 million parameters, respectively. This categorization makes the survey of the massive literature on semantic segmentation concise and easily tractable. 

\subsection{Large-scale networks}
The first end-to-end trainable network in this field was proposed by \cite{long2015fully}. They adapted the famous classification networks AlexNet \cite{krizhevsky2012imagenet}, GoogleNet \cite{szegedy2014going}, VGG Net \cite{simonyan2014very} and through transfer learning, fine-tuned these networks for semantic segmentation. Reference \cite{badrinarayanan2017segnet} employed a symmetric encoder-decoder network where the upsampling of the low-resolution feature maps from encoder was done using the pooling indices. To increase the receptive fields of the convolutional kernals, dilated convolutions were used by the authors of \cite{chen2017deeplab}. This enabled them to harvest more contextual information that is very conducive to accurate segmentation. To capture the contextual information at multiple scales, \cite{zhao2017pyramid} proposed a Pyramid Pooling Module (PPM). Various variants of PPM are still being used to harvest contextual informations \cite{hong2021deep}. Based on the PPM module and ResNet \cite{he2016deep} as the backbone model, they proposed PSPNet. To preserve the finer details that are lost due to downsampling operations like strided convolution and pooling operations, \cite{lin2017refinenet} proposed RefineNet. They demonstrated that features from all the layers of a network are crucial so they designed their network such that it refines low resolution (coarse) features from deep layers using high-resolution (fine-grained) features from shallow layers. PSPNet was adopted by the authors of ICNet \cite{zhao2018icnet} as their backbone network. The PSPNet is incorporated in the primary branch which operates at very low resolution and two additional shallow branches are used to harvest finer spatial details. BiSeNet \cite{yu2021bisenet} uses two-branch approach; a deep branch for semantic information and a shallow one for spatial details. The output from these two branches then get integrated to give finely delineated prediction maps. HyperSeg \cite{nirkin2021hyperseg} uses EfficientNet \cite{tan2019efficientnet} as its backbone architecture. They also use weight prediction network to progressively upsample the low-resolution feature maps from the context head module. The decoder receives feature maps from different stages of the encoder and fuse them in a step-by-step fashion.
 
 \subsection{Mid-scale networks}
  ERFNet \cite{romera2017erfnet} used factorization strategy and applied it to the conventional convolutional kernals to decompose a $3 \times 3$ kernal into a pair of $3 \times 1$ and $1 \times 3$ 1-dimensional asymmetrical convolutions. MaskNet \cite{atif2022semantic} adopts the ERFNet and uses a shallow parallel branch to apply semantic masking technique to improve the accuracy of small classes. BiAttenNet \cite{li2021biattnnet} adopts a two-branched approach for the separation of the attention blocks into the spatial detailing branch to explore details in a specialized manner. They also employed FCN style ResNet to achieve rough segmentation. Later on, they merge the coarse branch and the detail branch to achieve a speed-accuracy balance. RegSeg \cite{gao2021rethink} designed a basic block which is inspired by ResNeXt \cite{xie2017aggregated} blocks and used two parallel dilated convolutional blocks to enhance the receptive-field while preserving low-level details also. Inspired by BiSeNet \cite{yu2018bisenet}, the authors of CABiNet \cite{kumaar2021cabinet} also proposed a two-branched methodology where global aggregation and local distribution blocks \cite{li2019global} have been employed to harvest long-range and short-range contextual informations that is key to achieving decent accuracy-efficiency balance.

 \begin{figure*}[t]
  \centering
  \includegraphics[width=0.9\linewidth]{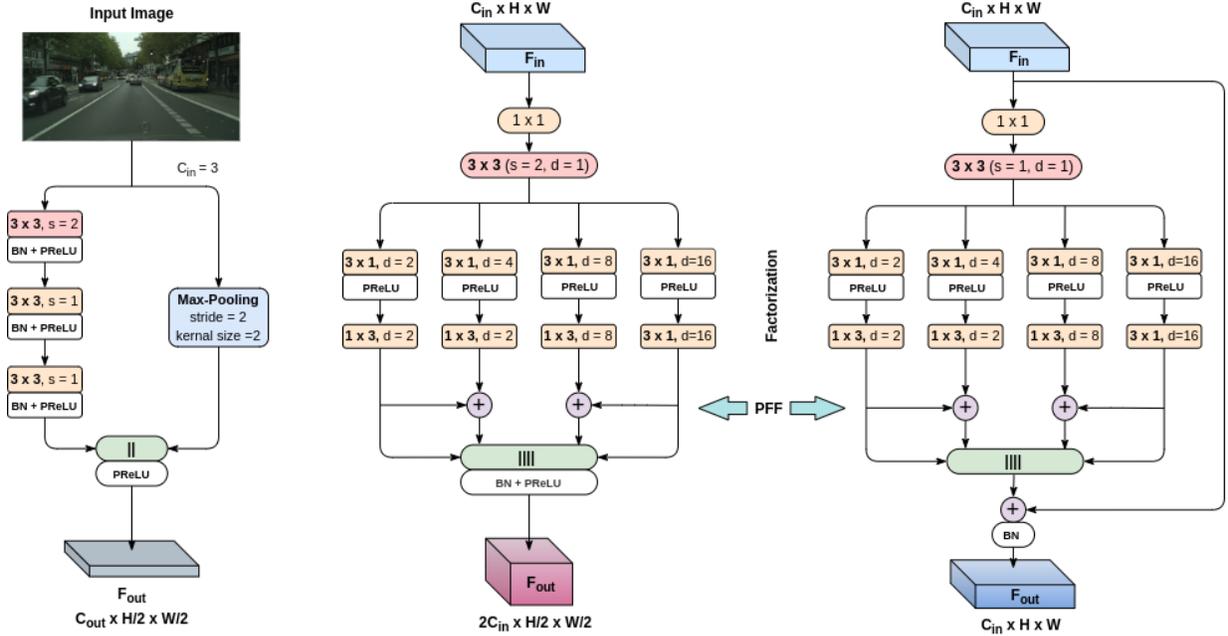}
  \caption{Basic building blocks of the proposed FPLNet. Left- Initial module. Centre- FPL-downsampler. Right- FPL module. BN and PReLU corresponds to Batch-normalization and Parametric recified linear unit, respectively.}
  \label{fpl_module}
\end{figure*}

 \subsection{Lightweight networks}
The authors of DABNet \cite{li2019dabnet} proposed a simple architecture to achieve a state-of-the-art performance in lightweight scenario. More specifically, they designed a simple depth-wise asymmetric bottleneck to process incoming stream of features. Each module consists of two depth-wise convolutions; one conventional and the other one dilated. Following the success of two-branched approach, ContextNet \cite{poudel2018contextnet},  fuses the low-resolution deep semantic feature with the high-resolution spatial finer details.  \\

\subsection{Ultra-lightweight networks}
With respect to accuracy oriented networks, ENet \cite{paszke2016enet} focuses the other extreme and consequently achieved a real-time network through aggressive parameter reduction. This work introduced a new direction of research, i.e., real-time methods. As a result, other researchers followed suit and many efficient networks were proposed. ESPNet \cite{mehta2018espnet} is one such efficient model that has less than 0.5 million parameters. It uses a bank of convolutions with different dilation rates, allowing it to capture mult-scale context. CGNet \cite{wu2020cgnet} in its CG module, uses 3 types of context aggregation-- surrounding context, joint feature and global context extractor. This is also an ultra-lightweight network with less than 0.5 million parameters suitable for resource-constrained devices.

\textbf{Gap in the literature:}
In summary, the large-scale models achieve great accuracies with networks having millions of parameters which are not suitable for mobile devices. Similarly, the mid-scale networks, thought smaller than large-scale counterparts, still have million of parameters prohibiting them from being a decent choice for physical deployments in practicle scenarios. On the other extreme of spectrum, accuracies of the ultra-lightweight networks are not sufficient enough for practical deployments especially in applications that involve risk. So, to fill this gap, ultra-lightweight networks with decent accuracies are required. In this work, the FPLNet has been proposed to fill this gap.
 
\section{FPLNet: Proposed Network} \label{proposed_network}

In this section, the basic building blocks of the proposed FPLNet will be discussed first. A comparison of conventional convolution block, ESP module and the proposed FPL module has also been presented along with their merits and demerits to give deeper insights. Finally, the overall architecture design of the FPLNet is presented which includes the attached decoder.

\subsection{Basic Building Blocks}
The basic building blocks of the proposed FPLNet include initial module, FPL module and FIR module which are discussed in detail in the subsequent sections.\\

\subsubsection{\textbf{Initial module}}
In real-time semantic segmentation networks, for reducing the number of parameters and computational cost, usually, the spatial resolution is hastily downsampled to quarter resolution by using two consecutive downsampling blocks \cite{mehta2018espnet, romera2017erfnet}. This strategy undoubtedly saves some parameters and convolution operations. However, this aggressive reduction in spatial resolution so early has a significant negative effect on the fine delineation of the object boundaries that ultimately leads to poor accuracy performance \cite{paszke2016enet}. This is because much of the finer spatial information, that could have been harvested in the first stage, is lost due to early downsampling. So, to avoid the loss of finer details, we use two successive convolutional layers (excluding downsampler) in stage-1. In this way, we are able to harvest better spatial information at half spatial resolution in the first stage, that yields better feature representation at the input of the second stage. \\
\subsubsection{\textbf{FPL module}} 
In order to correctly recognize a small region of an image, we need the information of surrounding regions as well \cite{wu2020cgnet}. So, context plays a very important role in vision problems and especially in semantic segmentation.
The simplest way to increase the context integration is to enlarge the receptive fields of kernals by increasing their sizes. So, researchers have used kernals of different sizes ranging from $3 \times 3$ to $11 \times 11$ \cite{krizhevsky2012imagenet}. However, this strategy is very inefficient as it leads to increase in the parameters exponentially. For example, a $3 \times 3$ conventional kernal has 9 parameters, whereas an $11 \times 11$ kernal has 121 parameters per channel. One simple solution to this problem is to use dilated convolutions \cite{holschneider1990real}. Dilation allows us to increase the receptive field without increasing the number of parameters. For example, a $3 \times 3$ dilated convolution with a dilation factor 16 spans a region of $33 \times 33$ of input feature map, whereas to span the same region, a conventional kernal would require 1089 parameters which is $121 \times$ more than its dilated counterpart. So, with this strategy one can save a huge number of parameters. However, in complex pixel-level dense prediction tasks, such as semantic segmentation, single-scale context is not sufficient and we need multi-scale context. \\
In literature, the most popular way to harvest multi-scale context is to employ spatial pyramid \cite{mehta2018espnet}. Spatial pyramid is nothing but multiscale context aggregation. This multiscale context can be intergrated either by pooling features with different window sizes \cite{zhao2017pyramid, chen2017deeplab, hong2021deep} or by employing a bank of convolutional filters with different dilation rates \cite{mehta2018espnet}. The former is used once in the network usually before the final prediction stage \cite{zhao2017pyramid} whereas the latter is usually distributed across the network and is built into the basic blocks themselves \cite{mehta2018espnet}. This eliminates the need to use a seperate pyramid module for context aggregation as the basic blocks themselves are well equipped to carry out the required task. Therefore, we adopted the latter approach, i.e., context aggregation as an in-built functionality of the basic blocks. More specifically, we adapted the ESP module of the classical ESPNet and reformulated it. The ESP module, in order to save parameters and computations, drastically squeezes the channel depth of the input feature; 1/5 to be more specific. This saves the parameters at the cost of information loss along the channel dimension.
\begin{figure}
  \centering
  \includegraphics[width=1\linewidth]{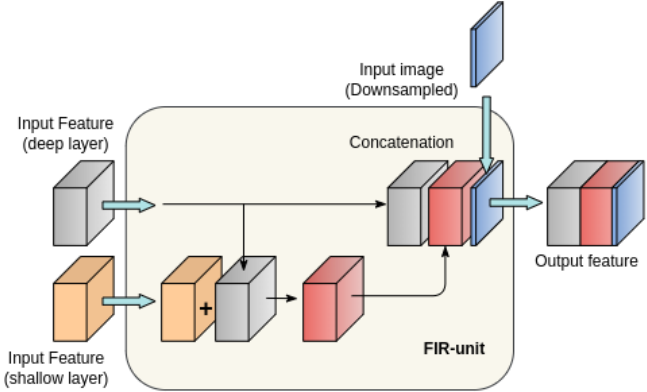}
  \caption{\small Feature-Image Reinforcement (FIR) unit: A dedicated unit to perform the fusion of shallow and deep feature tensors with the downsampled input image at the output of each stage.}
  \label{fir}
\end{figure}

The proposed approach, however, is distinct from \cite{mehta2018espnet} in two important ways. Firstly, we decomposed the spatial pyramid into two consecutive stage. We employed a conventional convolution in the first stage to gather local information whereas a bank of dilated convolutions is employed in the second stage to gather multi-scale context. Secondly, we factorized the spatial pyramid (i.e., bank of dilated convolutions) in the second stage which leads to significant parameter saving while incurring a slight toll on the accuracy. To be more specific, each $3 \times 3$ dilated kernal is factorized into a pair of $3 \times 1$ and $1 \times 3$ kernal. In this way, we are able to save three parameters in each channel of the kernal. The factorization also allows us to put an extra layer of non-linearity in between the two asymmetric kernals. This adds an additional linear region in the manifold space which leads to better accuracy with a negligible increment in computation cost \cite{romera2017erfnet}. Being on top of the factorized spatial pyramid of kernals, the conventional convolution in the first stage adds an additional layer of transformation which increases the effective depth of the network without increasing the number of basic modules. Moreover, being a symmetric convolution, it helps in balancing the accuracy loss incurred by the bank of factorized asymmetric 1D convolutional filters. This two-stage pyramid also enhances the effective receptive field with respect to ESP module \cite{szegedy2016rethinking}. More importantly, the pyramid decomposition enables us to solve the problem of gridding artifact in a quite simple way. To be more specific, we applied a simple pairwise feature fusion (PFF) that adds only two features as opposed to HFF of ESP module which engages all the five features of the module.
\begin{table}[h]
  \centering \makegapedcells
  \small
  \caption{Comparison of different modules. FPL-decomp. is the decomposed (two-stage) version of FPL without factorization of spatial pyramid in the second stage.}
  \begin{tabular}{l|c|r}
    \hline
     \textbf{Modules} & \textbf{Parameters} & \textbf{Numbers} \tstrut \bstrut \\
     \hline
     Convolution & $C_iC_ok^2$ & 32, 400 \tstrut \\ 
     ESP-original  & $\dfrac{C_o}{b} (C_i+k^2C_o)$ & 7, 200 \bstrut \\
     FPL-decomp. & $\dfrac{C_o}{b^2} (bC_i+k^2C_o+k^2bC_o)$  & 11, 025  \bstrut \\
     FPL proposed & $\dfrac{C_o}{b^2} (bC_i+k^2C_o+2kbC_o)$ & 8, 325 \bstrut \\ %\addlinespace
   \hline  
  \end{tabular}
  \label{param_comparison}
\end{table}

Table \ref{param_comparison} shows the comparison of different modules. $C_i$ is the number of channels in the incoming feature map and $C_o$ is the number of channels in the outgoing one. $k \times k$ is the size of the symmetric convolutional kernal. $b$ is the number of branches the incoming feature tensor is split into. The exact number of parameters per module, shown in Table \ref{param_comparison}, is computed for $C_i=C_o=60$, $k=3$ and $b=5$ for ESP-original and $b=4$ for both FPL-decomp. and FPL. Please note that the decomposition of pyramid both in FPL-decomp. and FPL results in enhanced receptive field (hence more wider context) and allows the depth of each branch to increase from 1/5 to 1/4. It is clear from the Table \ref{param_comparison} that the FPL-decomp. module requires 53.125 \% more parameters in comparison to ESP-original module. Our proposed final version of FPL module, on the other hand, achieves the same objectives as that of FPL-decomp. but with only 15.6 \% more parameters while having only a slight effect on the final accuracy.\\

\begin{figure*}[t]
\centering
  \includegraphics[width=1\linewidth]{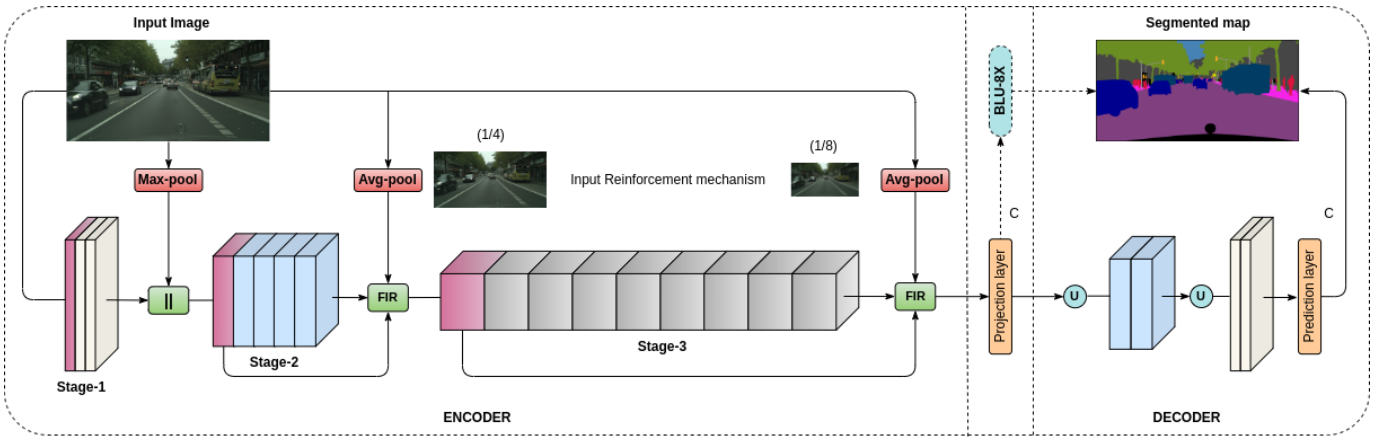}
  \caption{The network architecture of the proposed FPLNet. Stage-1 corresponds to the initial module having three convolutional layers. Stage-2 and stage-3 has 4 and 8 FPL modules, respectively. The first block of each stage is a downsampler, shown in pink color. The small circular units marked 'U' represent upsampling blocks which in our case is carried out by deconvolution operation followed by batch-normalization and PReLU activation function. ``BLU-8X": Bilinear-interpolation by a factor of 8.}
  \label{fplnet}
\end{figure*}

\subsubsection{\textbf{Feature-Image Reinforcement Unit (FIR unit)}}
Intra-stage skip connections are used to enhance the information flow of the network \cite{lo2019efficient, mehta2018espnet}. Image insertion in also done to improve feature representation at different stages of the encoder \cite{li2019dabnet}. To accomplish this, additive or concatenative fusion is usually used. Concatenation increases the parameters but performs slightly better than additive fusion. In this work, to improve it further, we design a dedicated FIR unit to jointly combine the deep and shallow features with the downsampled images. The proposed FIR unit employs a hierarchical combination of additive and concatenative fusion. This gives significantly better performance compared to addition or concatenation alone with the same number of trainable parameters. The FIR unit is shown in the Fig. \ref{fir}. Features from deep and shallow layers are first added. The resultant feature tensor is then concatenated with the feature from the deep layer follwed by image insertion.

\subsection{Network Architecture}
The overall network architecture of the FPLNet is shown in Fig. \ref{fplnet}. A stage of a network is that section which has continuous group of feature maps with same spatial resolutions. In this way, the encoder of the proposed network is composed of three stages; stage-1, stage-2 and stage-3. Stage-1 is instantiated by the initial module which has 2 convolutional layers for finer feature extraction. There are 4 and 8 modules in the second and third stage, respectively, excluding the modules acting as downsamplers. The downsampling operation of feature maps is employed in deep CNNs for two reasons. Firstly, it allows the deeper layers to receive rich contextual information which in turn provides better semantic information. Secondly, it reduces the computational cost by reducing the spatial dimensions of the feature maps. However, these two benefits are achieved at the cost of spatial information loss at the local level that is crucial to recover finer details in the final predicted map. Pooling operations such as max-pooling or average pooling are generally used for downsampling \cite{krizhevsky2012imagenet, badrinarayanan2017segnet} but we have used the same FPL module for downsampling as well simply by using strided convolution. This way we do not have to rely on pooling operations separately. The details of the architecture design is presented in Table \ref{arch_table}.

\begin{table}[h]
  \centering
  \small
  \caption{Detailed structure of the proposed network (FPLNet). ``Out Ch.": Number of feature channels at the Layer's output. ``Out Res." Output spatial resolution of feature tensors for input resolution $512 \times 1024$.}
  \begin{tabular}{l|c|lcc}
    \hline
      & \textbf{Layer} & \textbf{Operation} & \textbf{Out Ch.} & \textbf{Out Res.} \tstrut \bstrut \\
     \hline
     \multirow{9}{*}{\rotatebox{90}{ENCODER}} & 1 & Downsample (Conv-3) & 32 & $256 \times 512$ \tstrut \bstrut \\ 
     & 2-3 & Conv-3 & 32 & $256 \times 512$ \bstrut \\
     & 4 & Concatenation & 35 & $256 \times 512$ \bstrut \\
     \cline{2-5}
     & 5 & Downsample (FPL) & 64 & $128 \times 256$ \tstrut \bstrut \\
     & 6-9 & 4$\times$ FPL module & 64 & $128 \times 256$ \bstrut \\
     & 10 & FIR unit & 131 & $128 \times 256$ \bstrut \\
     \cline{2-5}
     & 11 & Downsample (FPL) & 128 & $64 \times 128$ \tstrut \bstrut \\
     & 12-19 & 8$\times$ FPL module & 128 & $64 \times 128$ \bstrut \\
     & 20 & FIR unit & 259 & $64 \times 128$ \bstrut \\
     \hline

     % \multirow{2}{*}{\rotatebox{90}{SWT.}} & 21 & Conv-1x1 & C & $64 \times 128$ \tstrut \\
     % & 22 & Bilinear interpolation & C & $512 \times 1024$ \bstrut \\
     % \hline

     \multirow{6}{*}{\rotatebox{90}{DECODER}} & 21 & Projection (Conv-1x1) & 64 & $64 \times 128$ \tstrut \bstrut \\
     & 22 & Upsampler & 64 & $128 \times 256$ \bstrut \\
     & 23-24 & 2$\times$ FPL & 64 & $128 \times 256$ \bstrut \\
     \cline{2-5}
     & 25 & Upsampler & 16 & $256 \times 512$ \tstrut \bstrut \\
     & 26-27 & 2$\times$ FPL & 16 & $256 \times 512$  \bstrut \\
     \cline{2-5}
     & 28 & Projection (Deconv) & C & $512 \times 1024$ \tstrut \\
   \hline  
  \end{tabular}
  \label{arch_table}
\end{table}

\begin{table*}[t]
  \centering
  %\resizebox{\columnwidth}{!}
  \caption{Evaluation results (per-class basis) of our model on cityscapes validation (Top) and test set (Bottom).}
  \scalebox{0.800}{
  %\makebox[0.70\textwidth]{
  \begin{tabular}{l|ccccccccccccccccccc|c}
  \hline
    \textbf{FPLNet} & \rot{\textbf{Road}} & \rot{\textbf{Sidewalk}} & \rot{\textbf{Building}} & \rot{\textbf{Wall}} & \rot{\textbf{Fence}} & \rot{\textbf{Pole}} & \rot{\textbf{Traffic light}} & \rot{\textbf{Traffic sign}} & \rot{\textbf{Vegetation}} & \rot{\textbf{Terrain}} & \rot{\textbf{Sky}} & \rot{\textbf{Pedestrian}} & \rot{\textbf{Rider}} & \rot{\textbf{Car}} & \rot{\textbf{Truck}} & \rot{\textbf{Bus}} & \rot{\textbf{Train}} & \rot{\textbf{Motorbike}} & \rot{\textbf{Bicycle}} & \rot{\textbf{mIoU (\%)}} \tstrut \bstrut \\ \hline
     
     Val. set & 94.46 & 76.42 & 87.37 & 54.77 & 52.55 & 50.09 & 46.70 & 62.43 & 89.04 & 58.92 & 88.96 & 65.36 & 45.73 & 88.71 & 63.11 & 75.79 & 66.29 & 42.25 & 62.68 & 66.93 \tstrut\\ 

     Test set & 95.93 & 75.95 & 88.75 & 60.32 & 57.65 & 51.41 & 49.15 & 61.69 & 89.28 & 54.65 & 92.02 & 62.19 & 37.40 & 89.13 & 68.88 & 74.56 & 63.54 & 31.50 & 55.36 & 66.28 \bstrut \\ 
    \hline
  \end{tabular}
  }
  \label{class_ious}
\end{table*}

The first block of each stage is a downsampling module as shown in Fig. \ref{fplnet} with pink color. The spatial resolutions of first, second and third stages are $1/2$, $1/4$ and $1/8$ of that of the input image. We have used the FIR units at the end of stage-2 and stage-3 to enrich the feature representation and to improve the flow of information. It should be noted here that this technique of input image insertion is very different than the target/label insertion \cite{yu2018bisenet, zhao2018icnet}. The objective of target/label insertion is to provide auxiliary loss mechanism at different depths of the network for enhanced learning. So, the target/label insertion mechanism is only used during the training phase and is absent while inference. The technique used in the proposed work, i.e., input image insertion using the FIR unit is used both during training and inference and hence does not rely on annotated images. 

To produce segmentation map from the encoder itself, its output is directly projected to C-dimensional space using a pointwise convolution. Then, a bilinear-interpolation layer is used to directly upsample the low-resolution map to high-resolution. This is depicted in a transition layer between encoder and decoder section as shown in Fig. \ref{fplnet}.  
To make the network lightweight, the decoders are either absent altogether \cite{wu2020cgnet, li2019dabnet} or they are designed to be extremely light-weight \cite{mehta2018espnet}. On the other hand, many model size agnostic networks, where achieving high accuracy is the main goal, employ very complex decoders \cite{nirkin2021hyperseg, rosas2021fassd}. Since, our objective is to find a balance between accuracy and model size, we employ a middle approach. More specifically, we design a small simple asymmetric decoder to progressively recover the finer spatial details. We empirically found that the encoder to decoder skip connections do not offer any accuracy gain in our case so we do not use such connections and keep our decoder purely sequential. Moreover, this sequential nature of the decoder makes it memory efficient especially during the training phase. 

\section{Experiments} \label{experiments}
In order to show the effectiveness of the proposed module and the corresponding network, extensive experiments have been carried out. General experimental setup, used to conduct the experiments, has also been presented along with the dataset and the evaluation metrics used to evaluate the performance of the proposed FPLNet. Most importantly, a comparison with other state-of-the-art methods have been presented to show the effectiveness of the propsosed methodology. A detailed ablation study of different design choices has also been presented.

\subsection{Dataset}
There are many road-scene datasets that are publicly available, for example, Cityscapes and CamVid \cite{cordts2016cityscapes, brostow2009semantic}. However, Cityscapes is one of the most widely used dataset. The images in this dataset are captured across 50 different cities with high-quality pixel-level annotations which makes it highly diverse. So, it has been used for training and evaluation of the proposed network. There are a total of 5000 finely annotated images, divided into training, testing and validation set having 2975, 1525 and 500 images, respectively. Test labels are not included in the dataset but can be accessed on an online test server. The images of this dataset are of very high resolution (2048 × 1024). Hence, usually the downscaled version (i.e., $1024 \times 512$) are used by the researches for training.

\subsection{Training details}
All the experiments in this work have been conducted on a Tesla V100 GPU using PyTorch framework with CUDA 10.2 and cuDNN backends. Mini-batch stochastic gradient descent (SGD) \cite{krizhevsky2017imagenet} is used as the optimizer for training the FPLNet with momentum 0.9 and weight decay $5 \times 10^{-4}$. We employed the ``poly" learning rate strategy which is given by
\[
lr = lr_{init} \times (1 - \frac{iter}{max\_iter})^{power}
\]
$lr_{init}$ corresponds to initial learning rate which is $4.5 \times 10^{-2}$ with power 0.9. The cross-entropy loss function is used as the loss function.  The training is done in two stages. The encoder part of the network is trained in the first stage with a batch size of 12 where the weights were initialized randomly. In the second stage, the decoder is attached to the pretrained encoder (trained in first stage). The batch size is 6 for training the whole network. Though, the original resolution of images are $1024 \times 2048$, we trained our network by sub-sampling original images by a factor of 2. We trained the encoder for 500 epochs and the complete network for 1000 epochs. For data augmentation, standard strategies such as horizontal flipping, cropping and scaling have been employed during training. Following \cite{paszke2016enet}, a class weighting scheme has also been used to mitigate the class-imbalance problem. As per this scheme, different weights are assigned to different classes during training; giving more weight to rare classes and less weight to dominant classes. To be more specific, each class is assigned the following weight:
\[
w_{class} = \frac{1}{ln(c+p_{class})}
\]
Where $p_{class}$ is the normalized frequency of the a class and $c$ is a constant that is set to 1.02.

\begin{table*}[t]
  \centering
  \small
  \caption{Comparison of our FPLNet with other state-of-the-art networks on Cityscapes dataset. Symbol "-" indicates that the corresponding values are not reported by the authors.}
  \begin{tabular}{llccccccc}
  \hline
  \multirow{2}{*}{\textbf{Methods}} & \multirow{2}{*}{\textbf{Backbone network}} & \multirow{2}{*}{\textbf{ImageNet pretrain}} & 
  \multicolumn{2}{c}{\textbf{mIoU} (\%)} & \multirow{2}{*}{\textbf{Parameters (M)}} & \multirow{2}{*}{\textbf{Resolution}} & \multirow{2}{*}{\textbf{Speed (FPS)}} \tstrut \\ 
  %\cline{4-5}
  & & & \textbf{Val} & \textbf{Test} & & & \tstrut \bstrut \\
  \hline 
  FCN-8s \cite{long2015fully} & VGG16 \cite{simonyan2014very} \ & \checkmark & - & 65.3 & 134.5 & - & - \tstrut \\  
  ICNet \cite{zhao2018icnet} & PSPNet50 \cite{zhao2017pyramid} & \checkmark & - & 69.5 & 26.5 & 2048 $\times$ 1024 & 30.3 \\
  SegNet \cite{badrinarayanan2017segnet} & self & \checkmark & - & 57 & 29.5 & 640 $\times$ 360 & 16.7 \\   
  BiSeNetV1 \cite{yu2018bisenet} & Xception39 \cite{chollet2017xception} & \checkmark & 69.0 & 68.4 & 5.8 & 1536 $\times$ 768 & 105.8 \\
  %BiSeNet2 \cite{yu2021bisenet} & Res18 & \checkmark & 74.8 & 74.7 & 49 & 1536 $\times$ 768 & 65.5 \\ 
  %BiSeNet-L \cite{yu2018bisenet} & self & \checkmark & 74.8 & 74.7 & 49 & 1024 $\times$ 512 & 65.5 \\
  %DDRNet \cite{hong2021deep} & self & \checkmark & 77.8 & 77.4 & 5.7 & 2048 $\times$ 1024 & 101.6 \\
  HyperSeg \cite{nirkin2021hyperseg} & EfficientNet-B1 \cite{tan2019efficientnet} & \checkmark & 78.2 & 78.1 & 10.2 & 1536 $\times$ 768 & 16.1 \bstrut\\
  HSBNet \cite{li2021hierarchical} & MobileNetV2 \cite{sandler2018mobilenetv2} & \checkmark & 73.1 & 73.1 & 12.1 & 2048 $\times$ 1024 & 32.2 \bstrut \\
  \hline
  ERFNet \cite{romera2017erfnet} & self & \ding{55} & 70 & 68 & 2.2 & 1024 $\times$ 512 & 41.7 \tstrut\\   
  CABiNet \cite{kumaar2021cabinet} & MobileNetV3 \cite{howard2019searching} & \checkmark & 76.6 & 75.9 & 2.64 & 2048 $\times$ 1024 & 76.5 \\  
  BiSeNetV2 \cite{yu2021bisenet} & self & \checkmark & 75.8 & 75.3 & 4.59 & 1536 $\times$ 768 & 47.3 \\ 
  RegSeg \cite{gao2021rethink} & self & \ding{55} & 78.13 & 78.3 & 3.34 & 2048 $\times$ 1024 & 30 \\  
  BiAttenNet \cite{li2021biattnnet} & ResNet-34 \cite{he2016deep} & \checkmark & 71.4 & 74.7 & 2.2 & 1024 $\times$ 512 & 89.2 \bstrut \\  
  \hline
  DABNet \cite{li2019dabnet} & self & \ding{55} & 69.1 & 70.1 & 0.75 & 2048 $\times$ 1024 & 27.7 \tstrut \\
  ContextNet \cite{poudel2018contextnet} & self & \ding{55} & 65.7 & 66.1 & 0.88 & 2048 $\times$ 1024 & 54  \bstrut \\
  EDANet \cite{lo2019efficient} & self & \ding{55} & - & 67.3 & 0.68 & - & 81.3 \bstrut \\
  \hline
  ENet \cite{paszke2016enet} & self & \ding{55} & - & 57 & 0.36 & 1024 $\times$ 512 & 74.9 \tstrut \\
  ESPNet \cite{mehta2018espnet} & self & \ding{55} & 61.4 & 60.3 & 0.36 & 1024 $\times$ 512 & 112 \\  
  CGNet \cite{wu2020cgnet} & self& \ding{55} & 63.5 & 64.8 & 0.5 & 2048 $\times$ 1024 & 17.6 \bstrut \\  
  \hline
  \textbf{FPLNet (Proposed)} & self & \ding{55} & 66.93 & 66.28 & 0.49 & 1024 $\times$ 512 & 95.5 \tstrut \bstrut \\ % 97 for 1024x512
  \hline
  \end{tabular}
  \label{main_table}
\end{table*}

\subsection{Performance analysis}
Mean Intersection over Union (mIoU) is the most commonly used metric in the field of semantic segmentation to evaluate the accuracy of the model \cite{long2015fully}. The Intersection over union (IoU) for a particular class is defined as the ratio of overlap and union between the class prediction and the class ground truth. In case of multi-class segmentation, the mIoU of the model is calculated by taking the IoU of each class and averaging them out over all the classes present in all the predicted images.
    \[IoU = \frac{TP}{TP+FP+FN}=
    \sum_{i=1}^{k}n_{ii}/(t_i + \sum_{j=1}^{k}n_{ji}-n_{ii})\]
    where TP, FP and FN are, respectively, the number of true positives, false positives and false negatives at pixel level.
Table \ref{class_ious} shows class-wise and mean IoUs for both validation and Test set.

\subsubsection{\textbf{Comparison of the proposed FPLNet with other state-of-the-art methods}}
%We have compared our network with a number of state-of-the-art networks in terms of three commonly used performance evaluation metrics-- mIoU (\%), parameters (M) and speed (FPS). 
The comparison of FPLNet with other state-of-the-art networks has been summerized in Table \ref{main_table}. To measure accuracy, mIoU (\%) has been used and for efficiency, we use parameters (M) and speed (FPS). Apart from that, other comparative indicators such as type of backbone network, pretraining information, etc., are also included to facilitate a multifaceted perspective.
To keep the discussion tractable, we divide the sementic segmentation models into 4 broad categories based on number of parameters as shown in Table \ref{model_categories}.

\begin{table}[h]
  \centering
  \small
  \caption{Categorization of models based on their corresponding sizes. The number of parameters are presented in millions. }
  \begin{tabular}{c|c|c|c}
    \hline
     \textbf{Large-scale} & \textbf{Mid-scale} & \textbf{Lightweight} & \textbf{Ultra-lightweight} \tstrut \bstrut \\
     \hline
    $\ge5$ & $1-5$ & $0.5-1$ & $\le0.5$ \tstrut \bstrut \\      
   \hline  
  \end{tabular}
  \label{model_categories}
\end{table}

Recent large-scale networks such as BiSeNetV1 \cite{yu2021bisenet}, HyperSeg \cite{nirkin2021hyperseg} and HSBNet \cite{li2021hierarchical} achieve excellent accuracies. However, to achieve such high accuracies, they had to design networks with huge number of parameters. More specifically, BiSeNetV1, HyperSeg and HSBNet achieve 68.4, 78.1 and 73.1\% mIoU at 5.8, 10.2 and 12.1 million parameters, respectively. It is quite interesting to note here that HyperSeg \cite{nirkin2021hyperseg} is 5\% more accurate than HSBNet despite having almost 2 millions less parameters. This shows that a straight-forward increase in parameters does not guarantee proportional accuracy boost. This can be further observed, when we compare mid-scale networks with the large-scale ones. When we Compare CABiNet \cite{kumaar2021cabinet} with BiSeNetV1, we find that CABiNet has 7.5\% more accuracy than BiSeNetV1 while being $2.2\times$ smaller. Similarly, RegSeg \cite{gao2021rethink} and BiAttenNet \cite{li2021biattnnet} are 0.2 and 1.6\% more accurate than HyperSeg \cite{nirkin2021hyperseg} and HSBNet \cite{li2021hierarchical}, while being $3\times$ and $5.5\times$ smaller, respectively. A similar effect can be seen when we compare BiSeNetV1 and BiSeNetV2 \cite{li2021hierarchical}. Despite having 1.21 million less parameters, BiSeNetV2 is almost 7\% more accurate than BiSeNetV1.
 These observations demonstrate the possibility for smaller networks of achieving accuracies similar to multiple-times bigger networks through smart architecture design. Exploiting this possibility is extremely important because model sizes of the large-scale and mid-scale networks become bottleneck in achieving real-time processing in resource-constrained devices \cite{kumaar2021cabinet}. To address this issue, many networks have been developed with less than 1 million parameters. 
 
\begin{figure}[t]
  \centering
  \includegraphics[width=\linewidth]{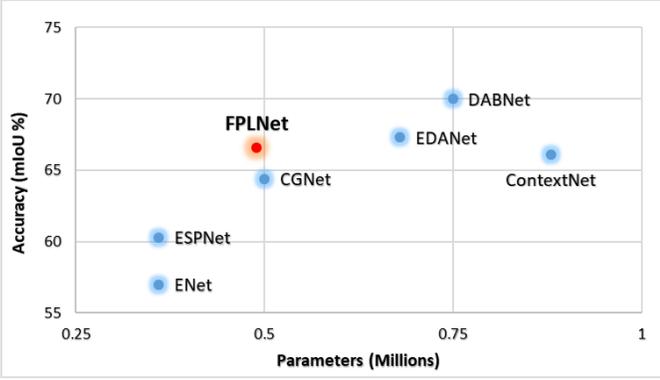}
  \caption{Mean IoU vs number of parameters of different lightweight state-of-the-art methods on cityscapes test set. With only less than 0.5 million parameters, the proposed FPLNet shown in red color achieves more than 66\% mIoU.}
  \label{scatter_plot}
\end{figure}

\begin{figure*}[h!]
\captionsetup[subfigure]{labelformat=empty}
\centering
%\captionsetup{justification=centering}
\subfloat[]{\includegraphics[width=0.22\linewidth]{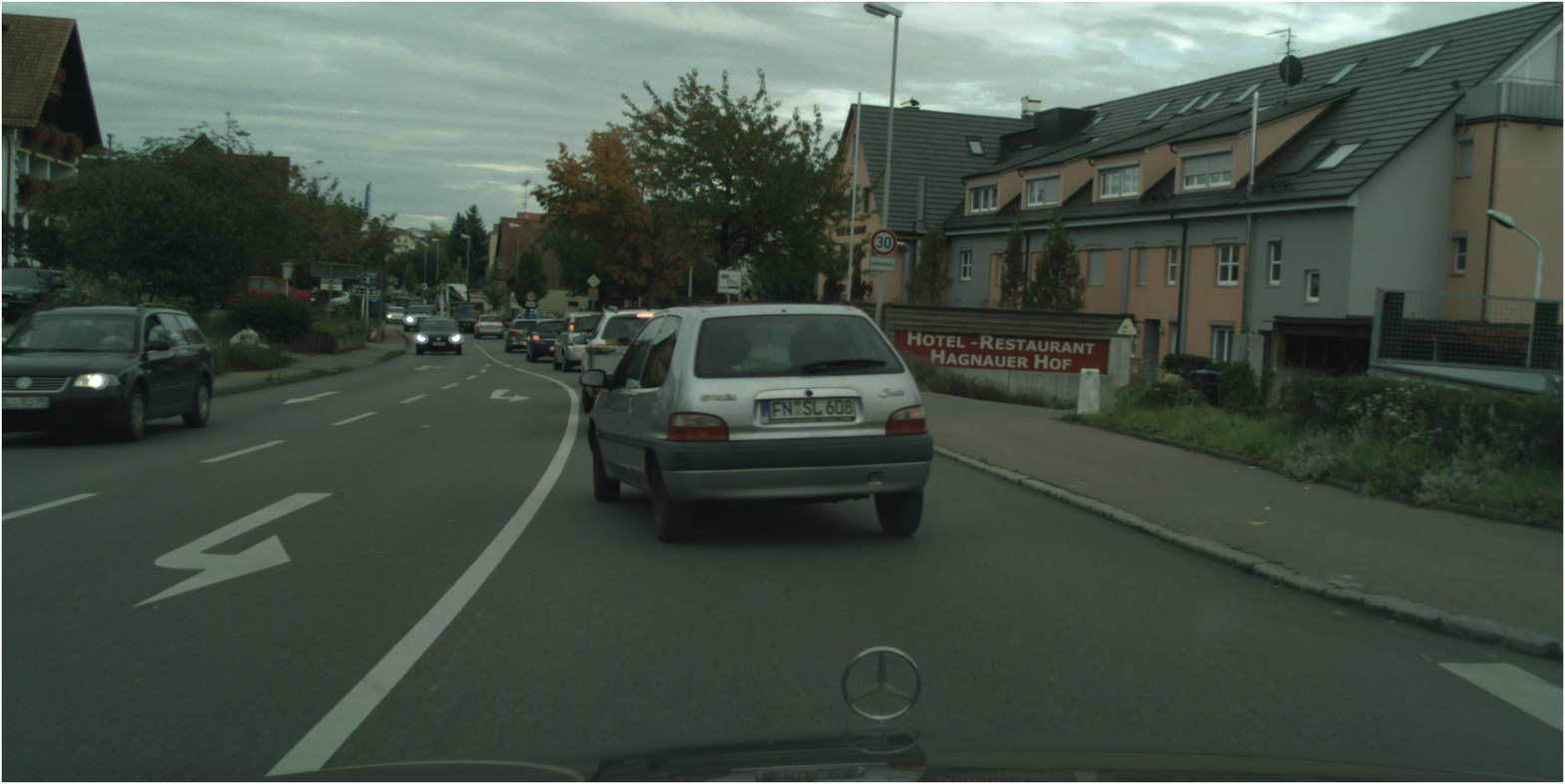}}\hspace{0.01cm}
\subfloat[]{\includegraphics[width=0.22\linewidth]{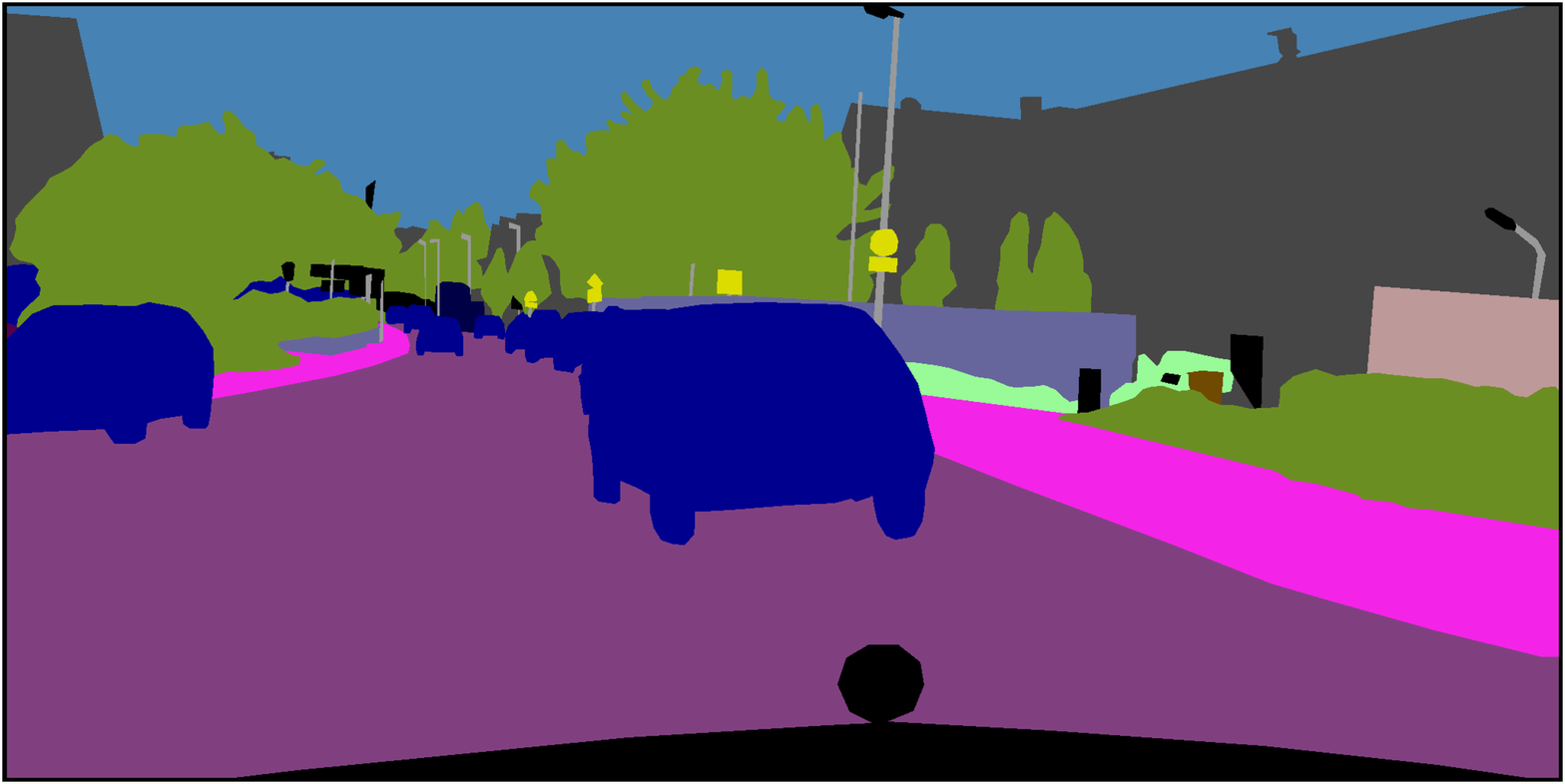}}\hspace{0.01cm}
\subfloat[]{\includegraphics[width=0.22\linewidth]{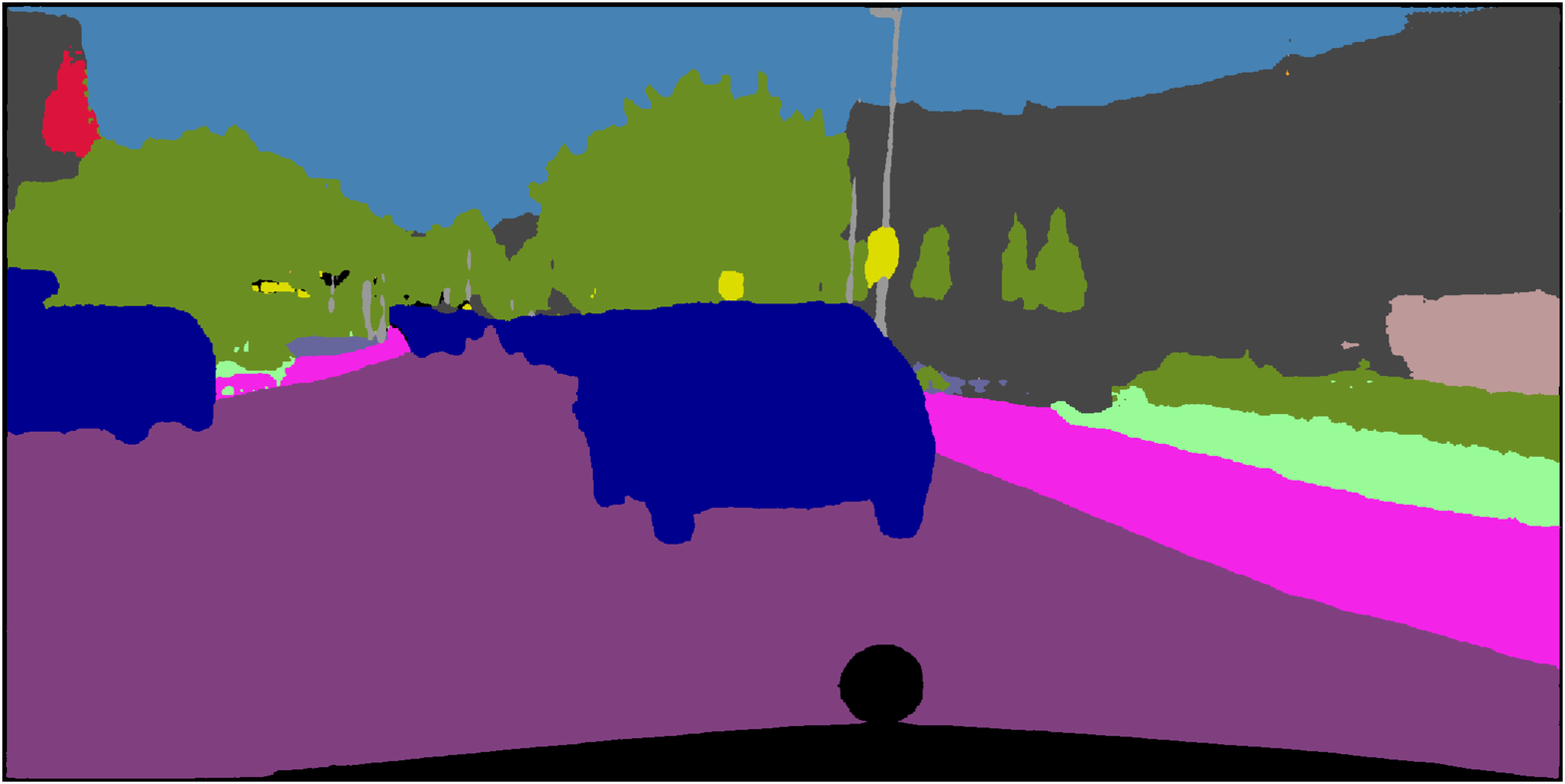}}\hspace{0.01cm}
\subfloat[]{\includegraphics[width=0.22\linewidth]{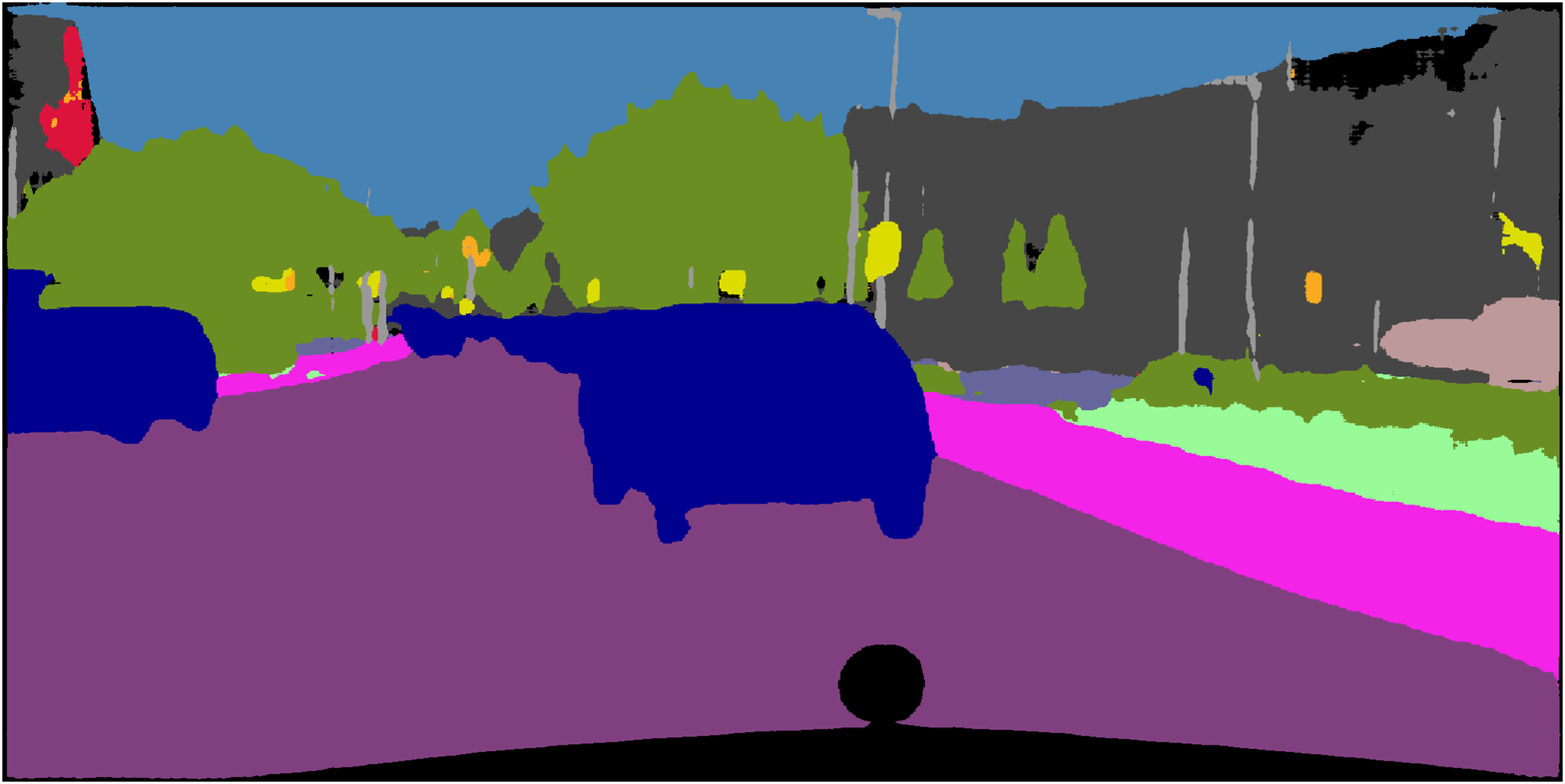}}\hspace{0.01cm}\\
\vspace{-0.75cm}
\subfloat[]{\includegraphics[width=0.22\linewidth]{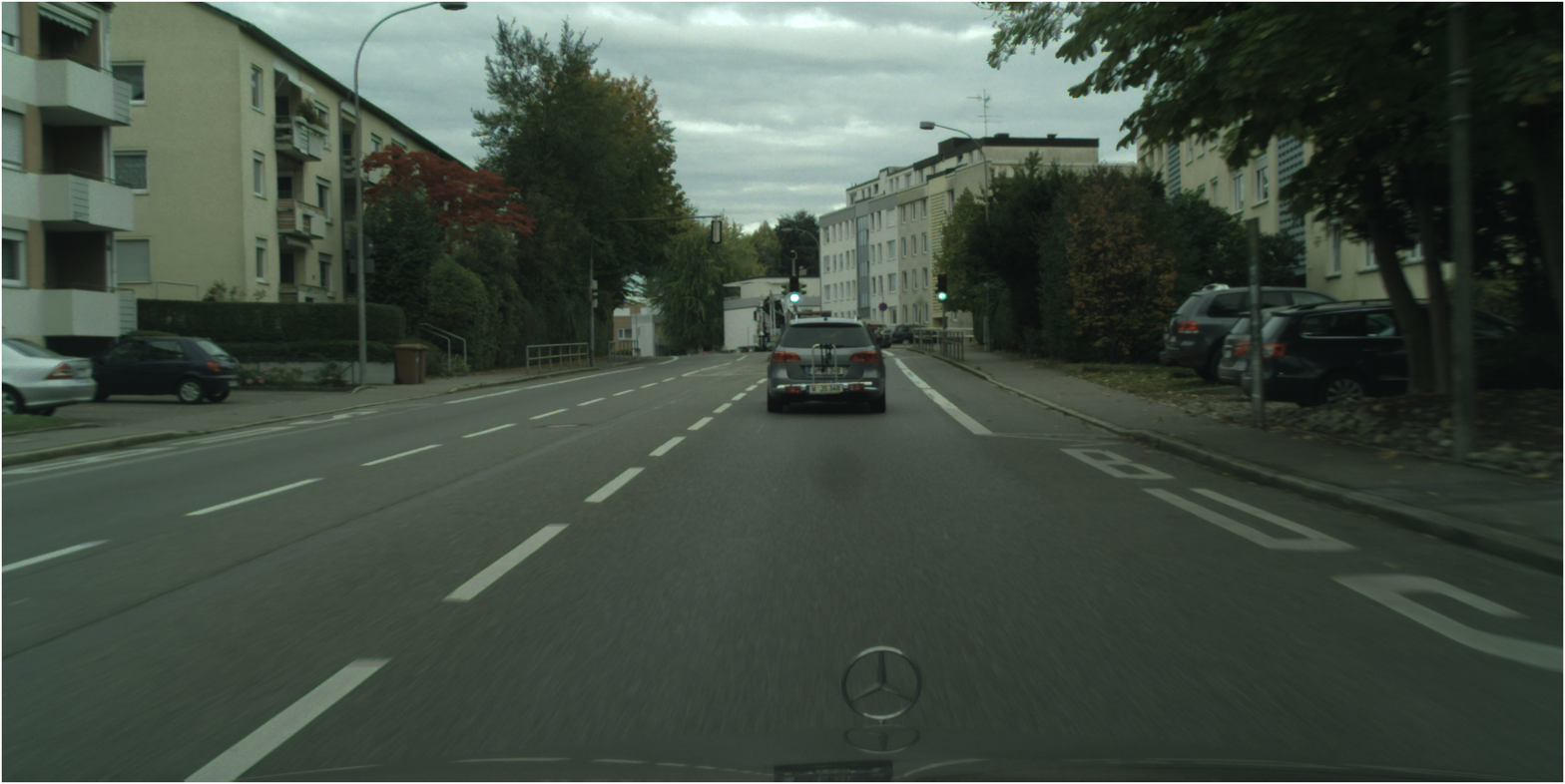}}\hspace{0.01cm}
\subfloat[]{\includegraphics[width=0.22\linewidth]{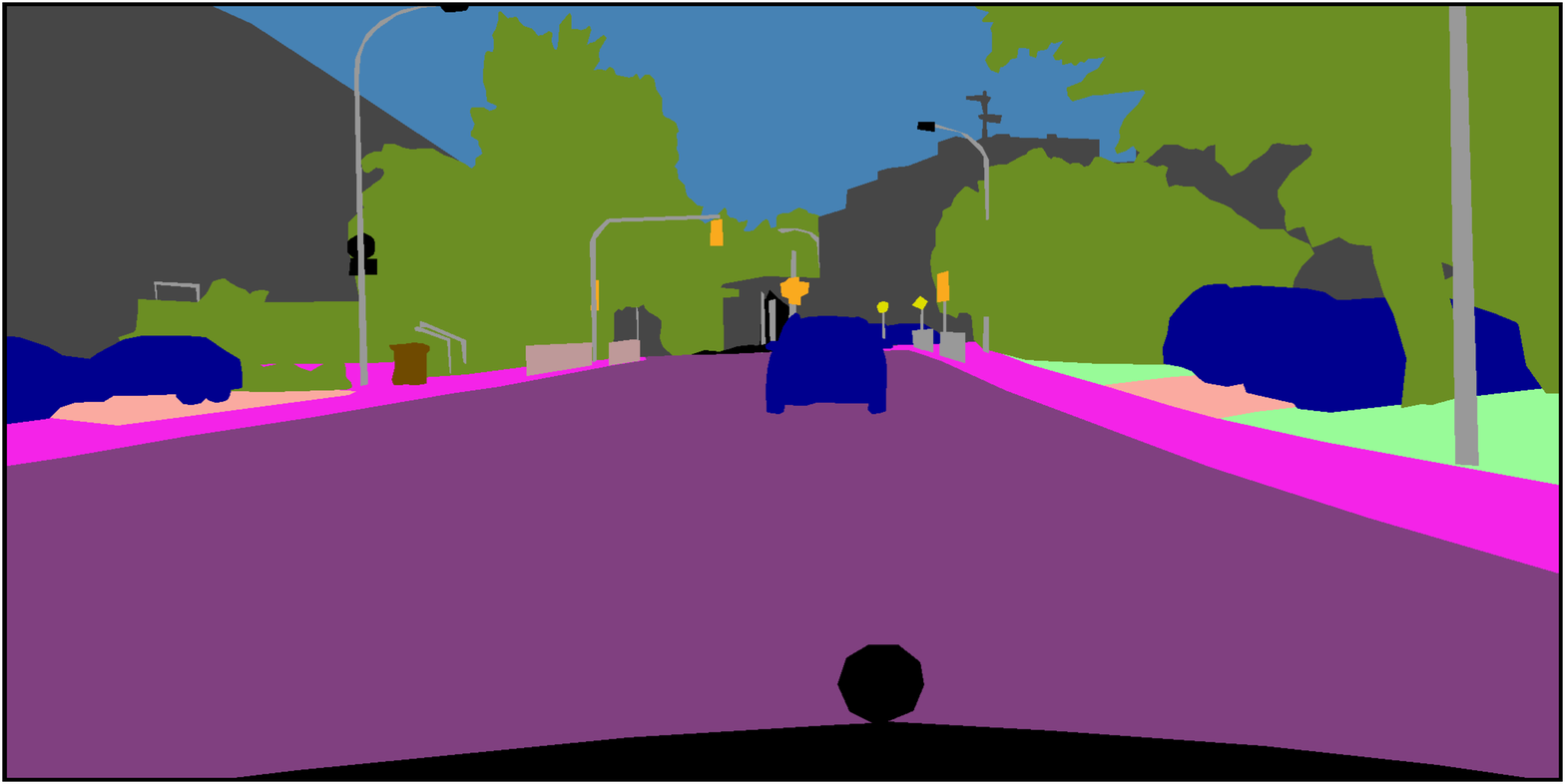}}\hspace{0.01cm}
\subfloat[]{\includegraphics[width=0.22\linewidth]{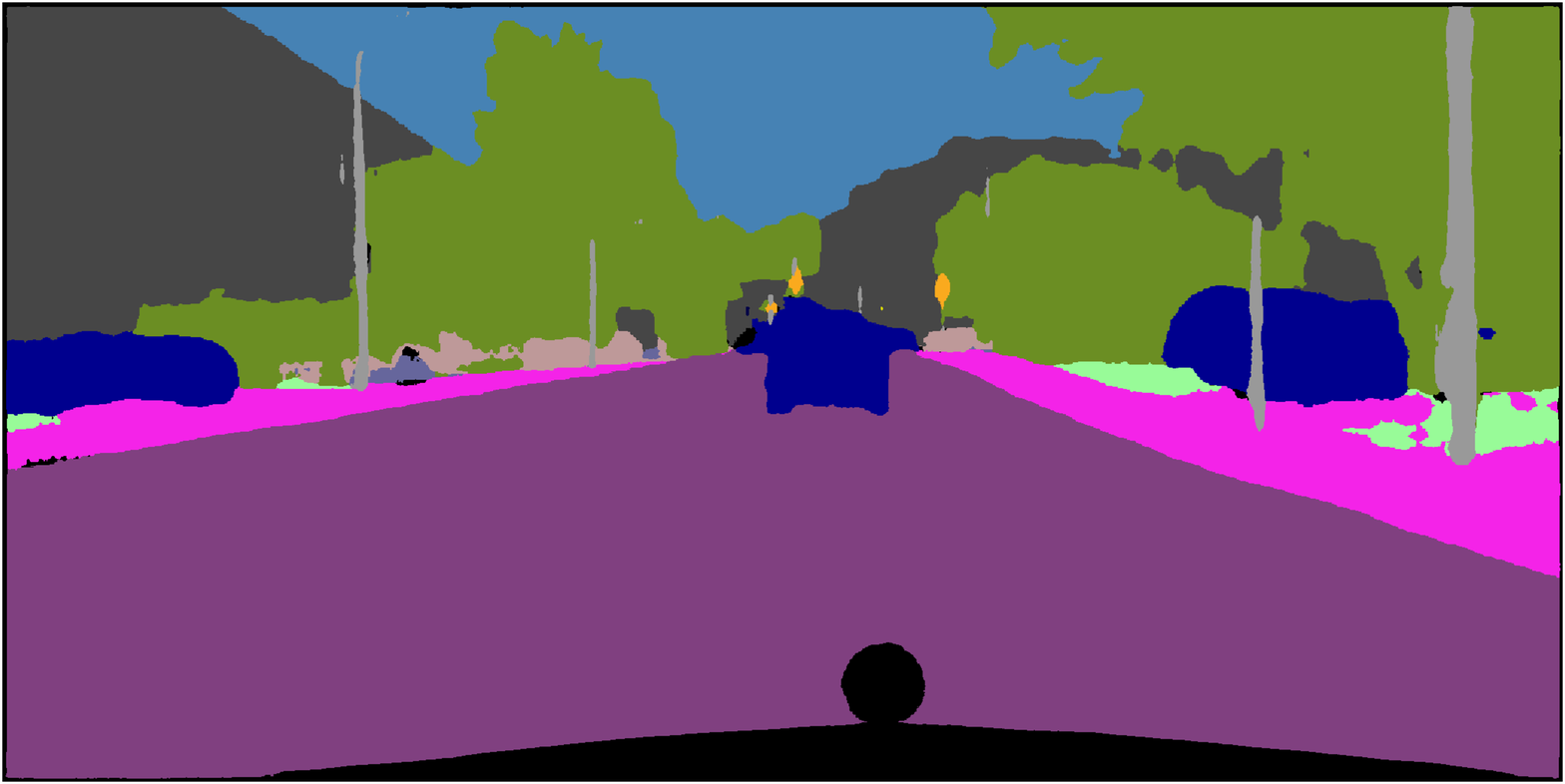}}\hspace{0.01cm}
\subfloat[]{\includegraphics[width=0.22\linewidth]{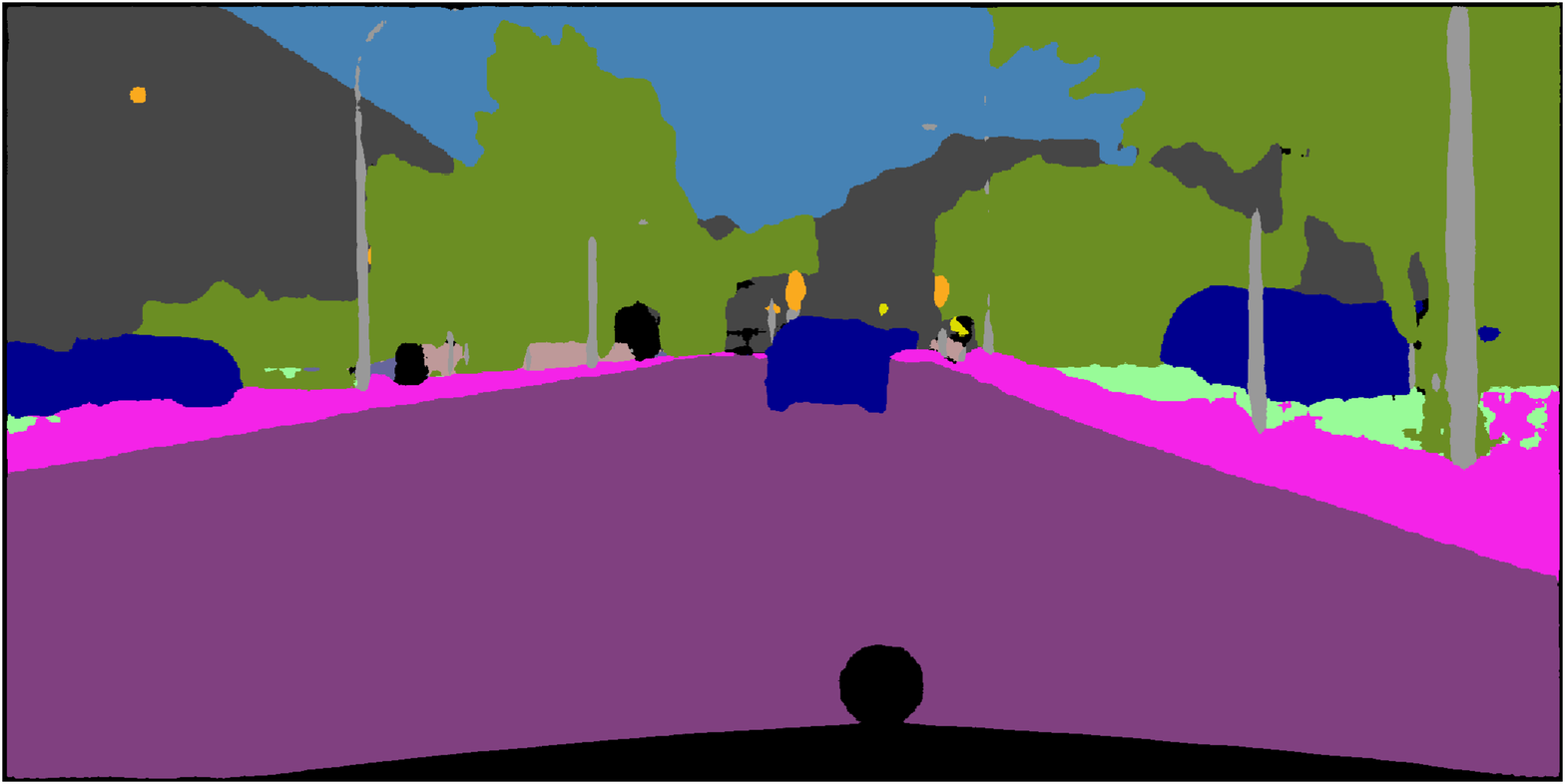}}\hspace{0.01cm}\\
\vspace{-0.75cm}
\subfloat[]{\includegraphics[width=0.22\linewidth]{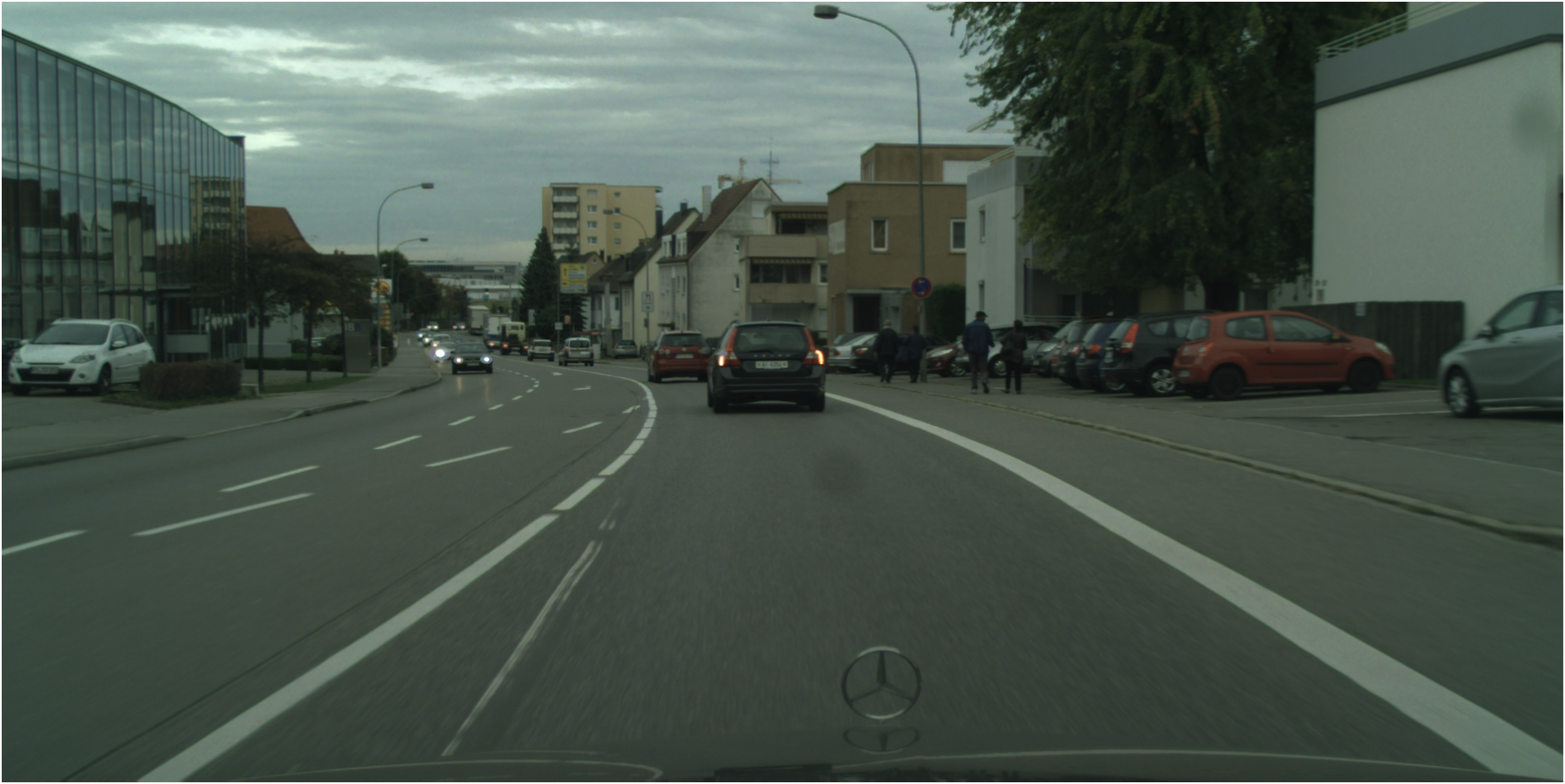}}\hspace{0.01cm}
\subfloat[]{\includegraphics[width=0.22\linewidth]{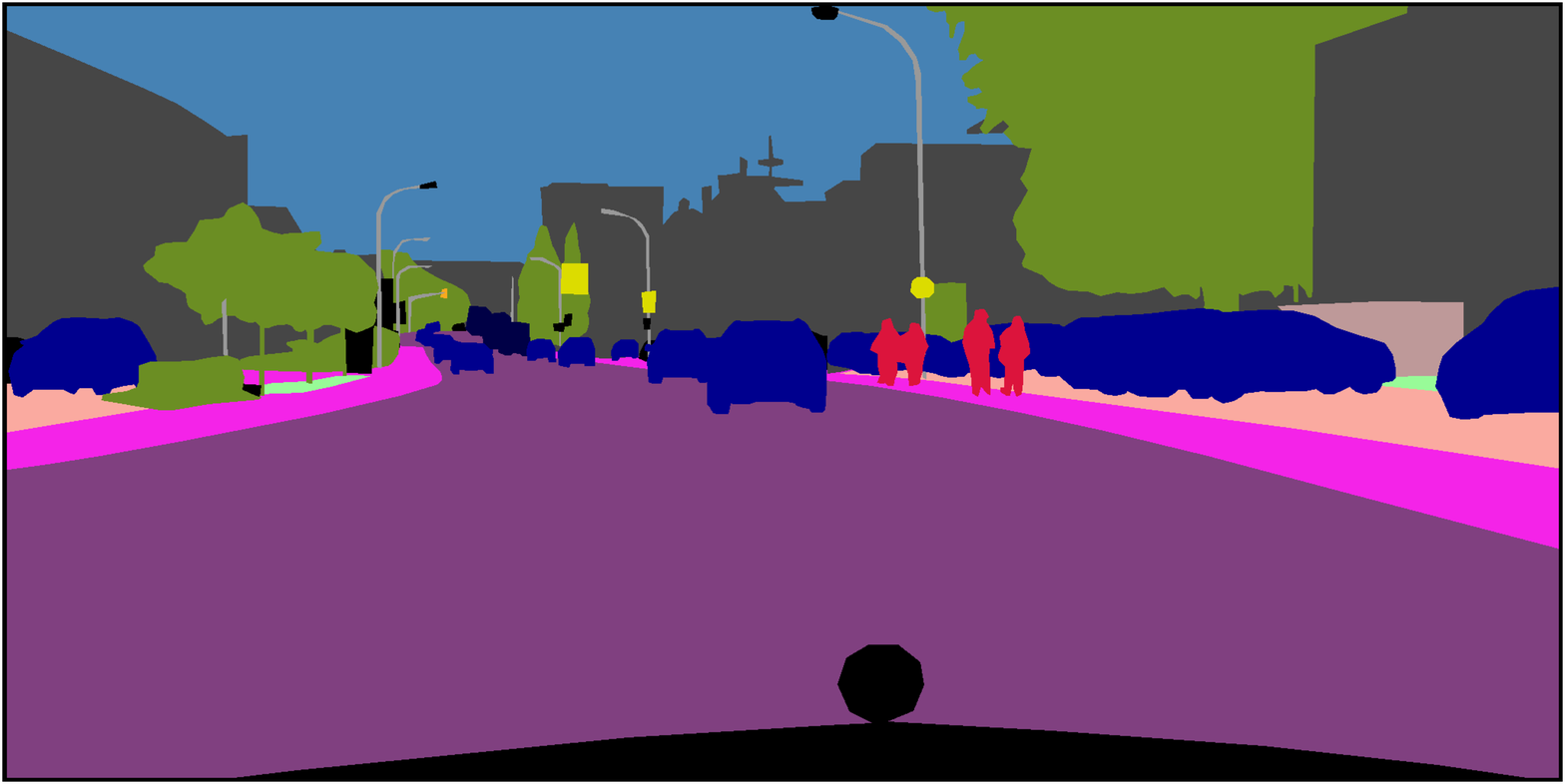}}\hspace{0.01cm}
\subfloat[]{\includegraphics[width=0.22\linewidth]{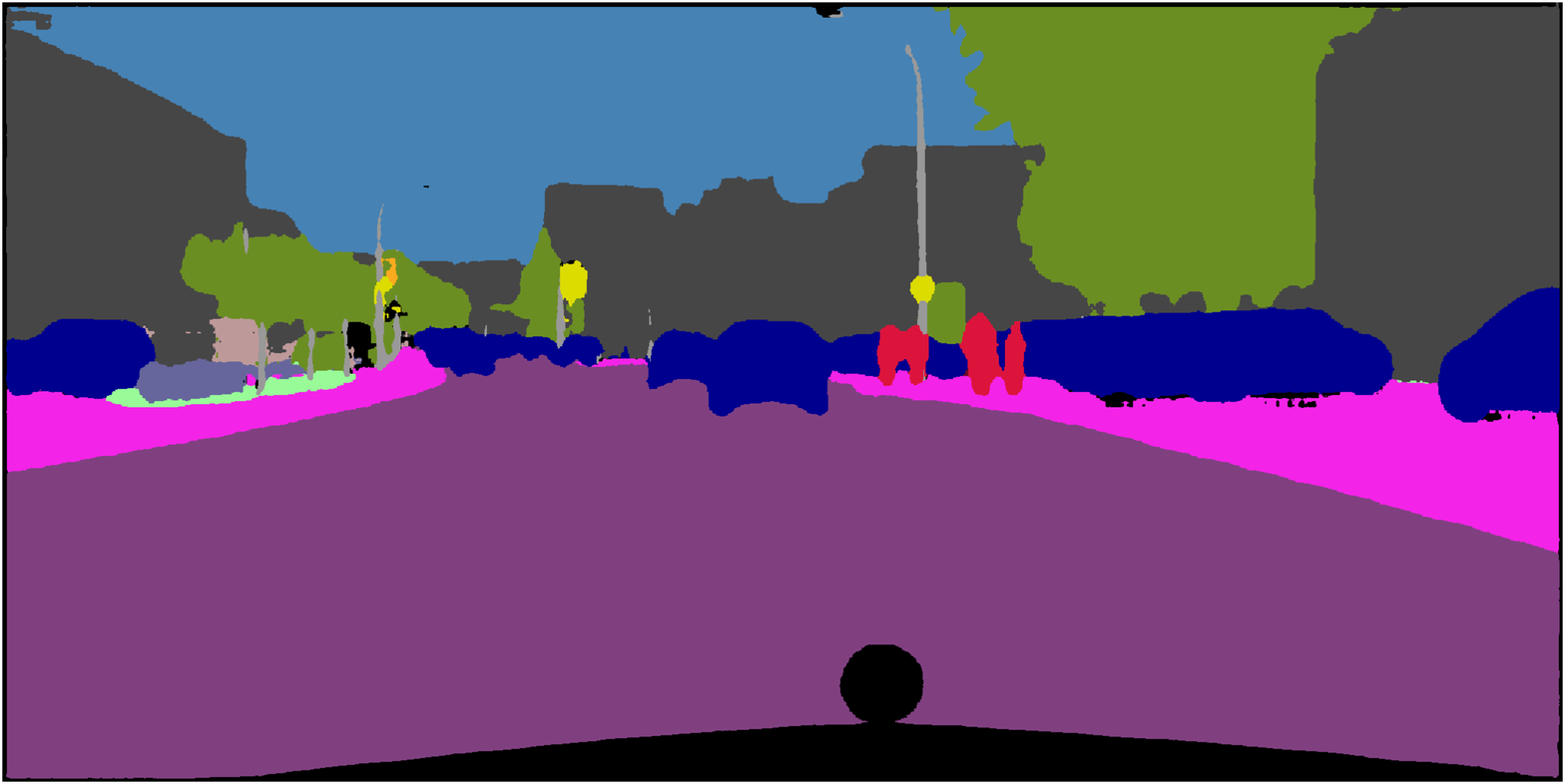}}\hspace{0.01cm}
\subfloat[]{\includegraphics[width=0.22\linewidth]{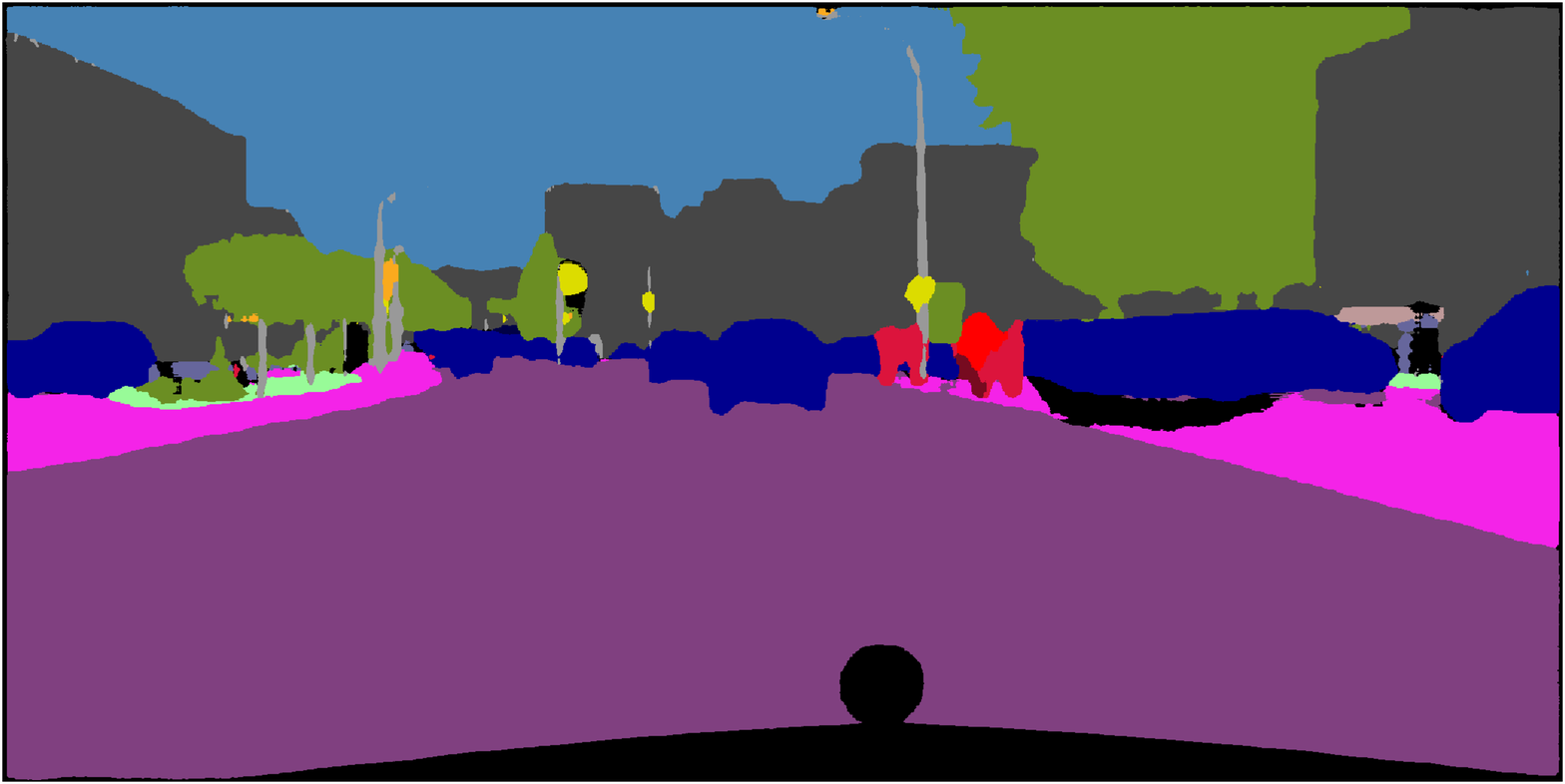}}\hspace{0.01cm}\\
\vspace{-0.75cm}
\subfloat[]{\includegraphics[width=0.22\linewidth]{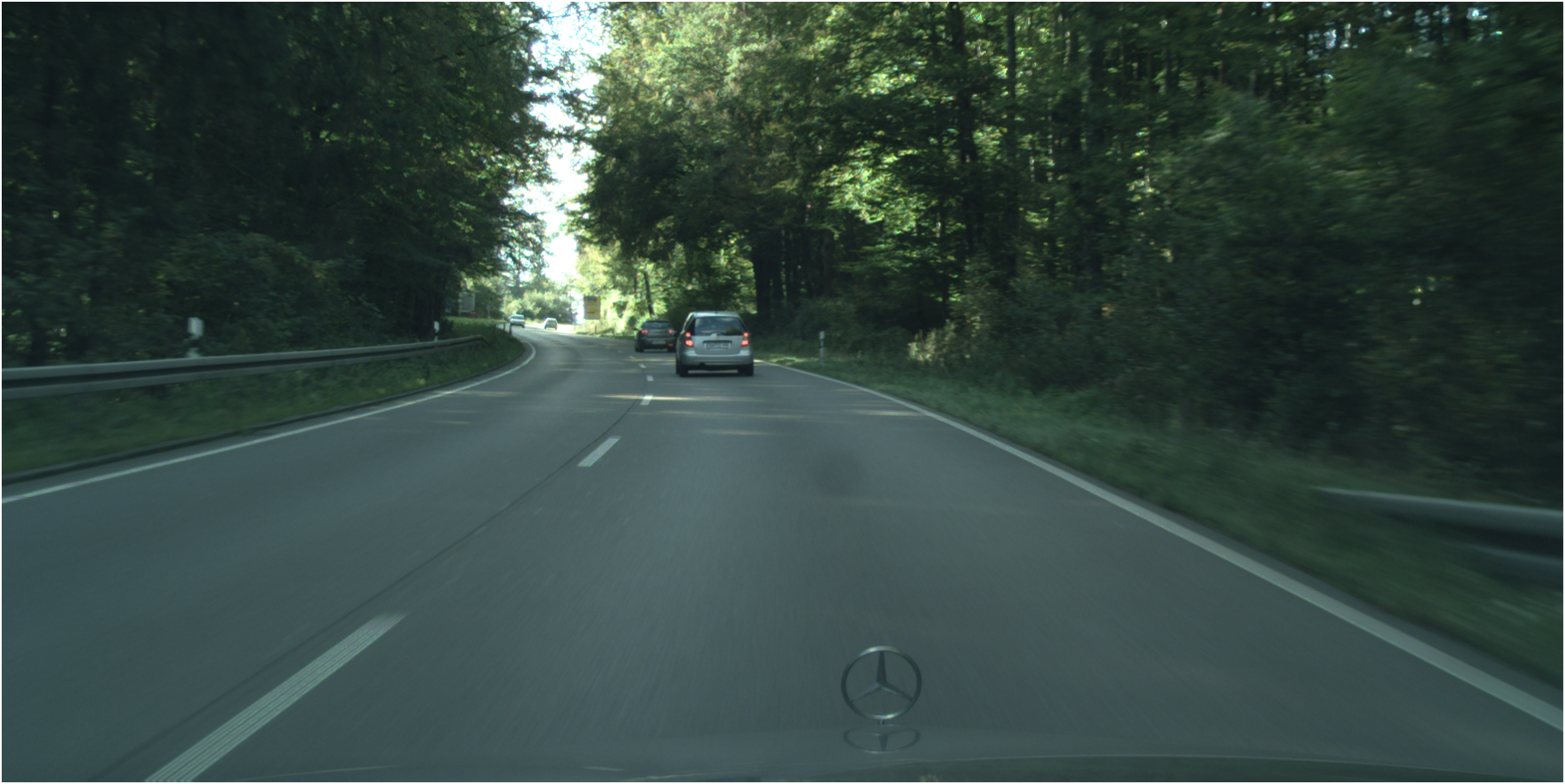}}\hspace{0.01cm}
\subfloat[]{\includegraphics[width=0.22\linewidth]{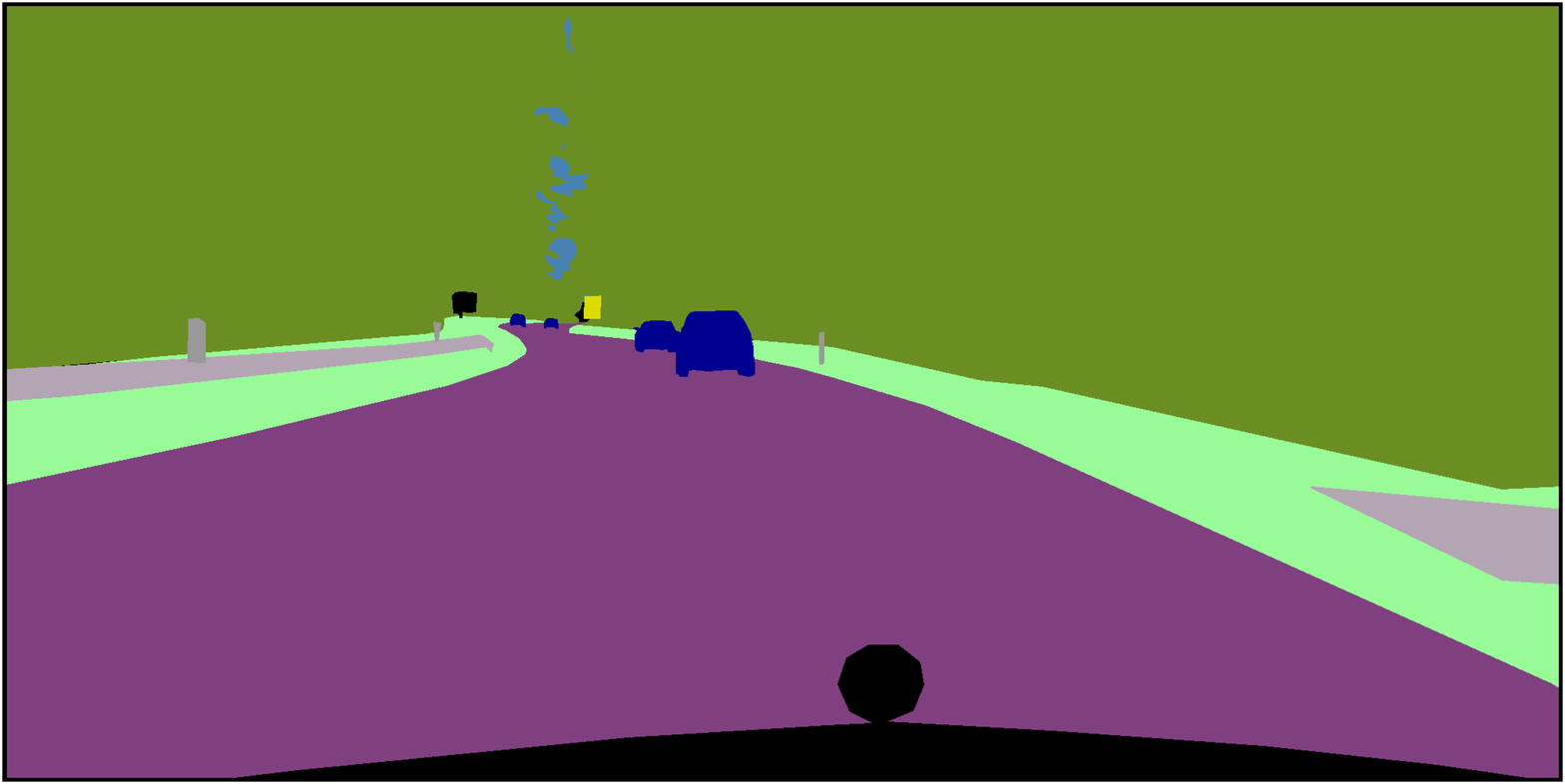}}\hspace{0.01cm}
\subfloat[]{\includegraphics[width=0.22\linewidth]{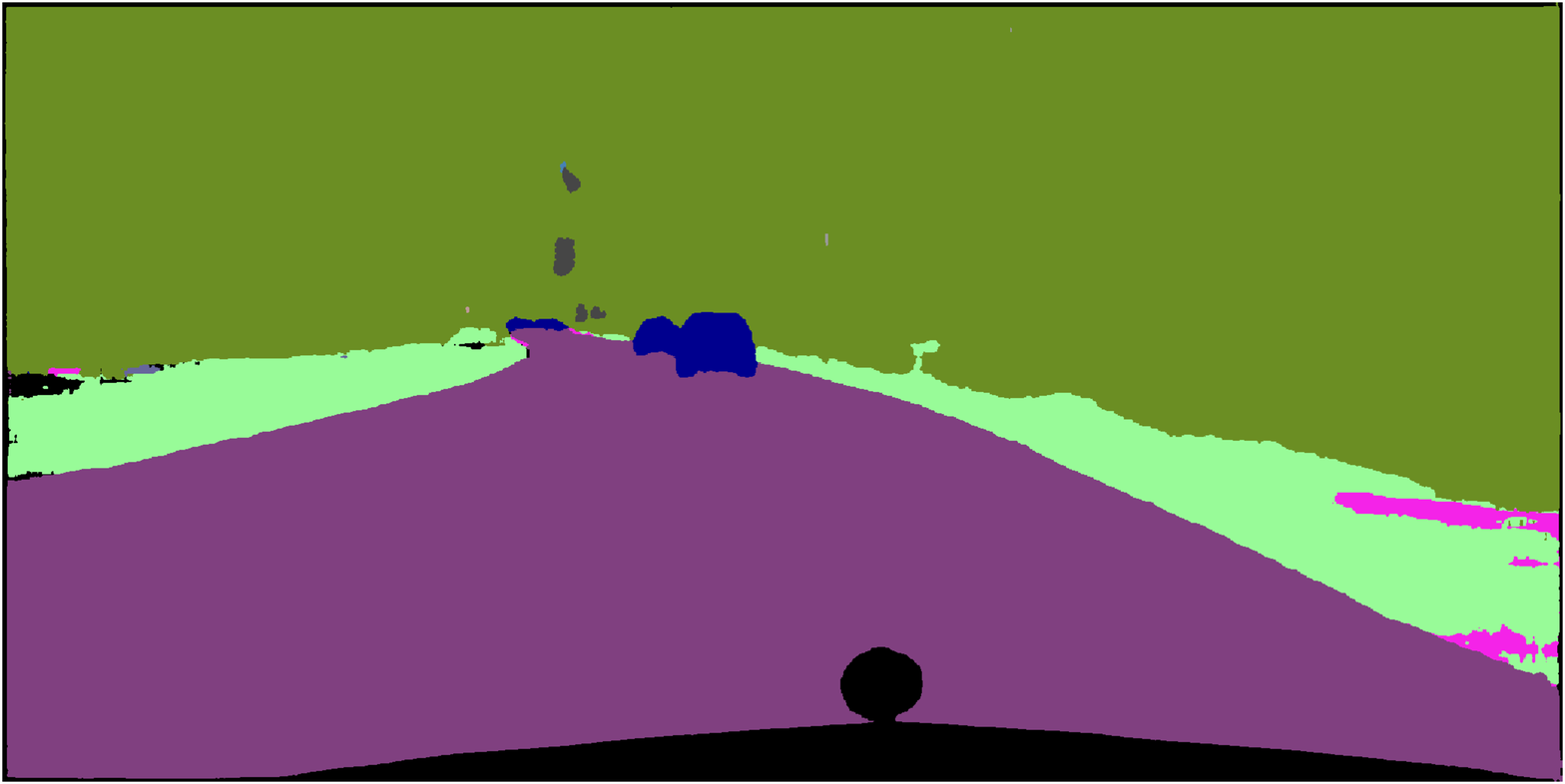}}\hspace{0.01cm}
\subfloat[]{\includegraphics[width=0.22\linewidth]{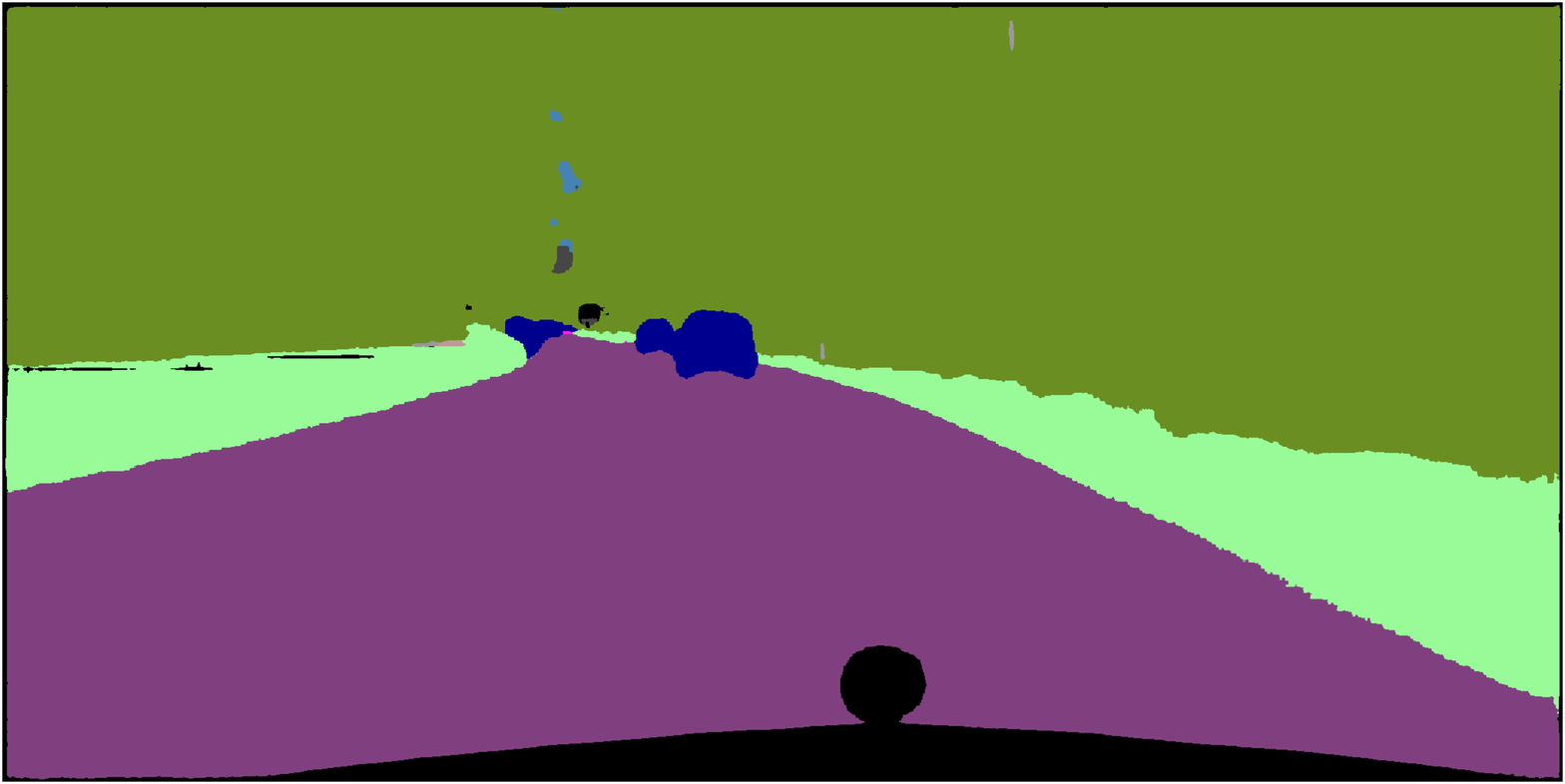}}\hspace{0.01cm}\\
\vspace{-0.75cm}
\subfloat[]{\includegraphics[width=0.22\linewidth]{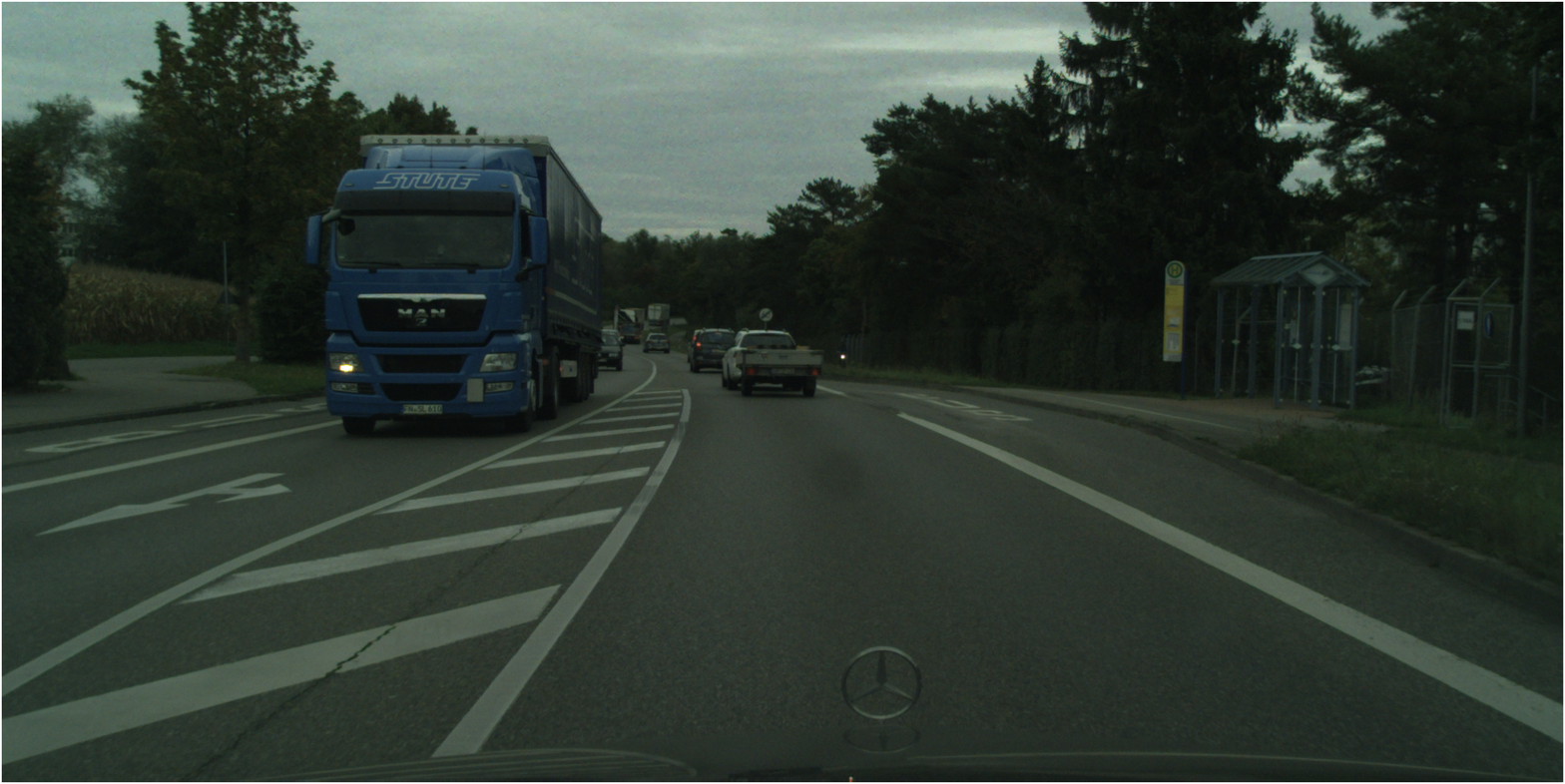}}\hspace{0.01cm}
\subfloat[]{\includegraphics[width=0.22\linewidth]{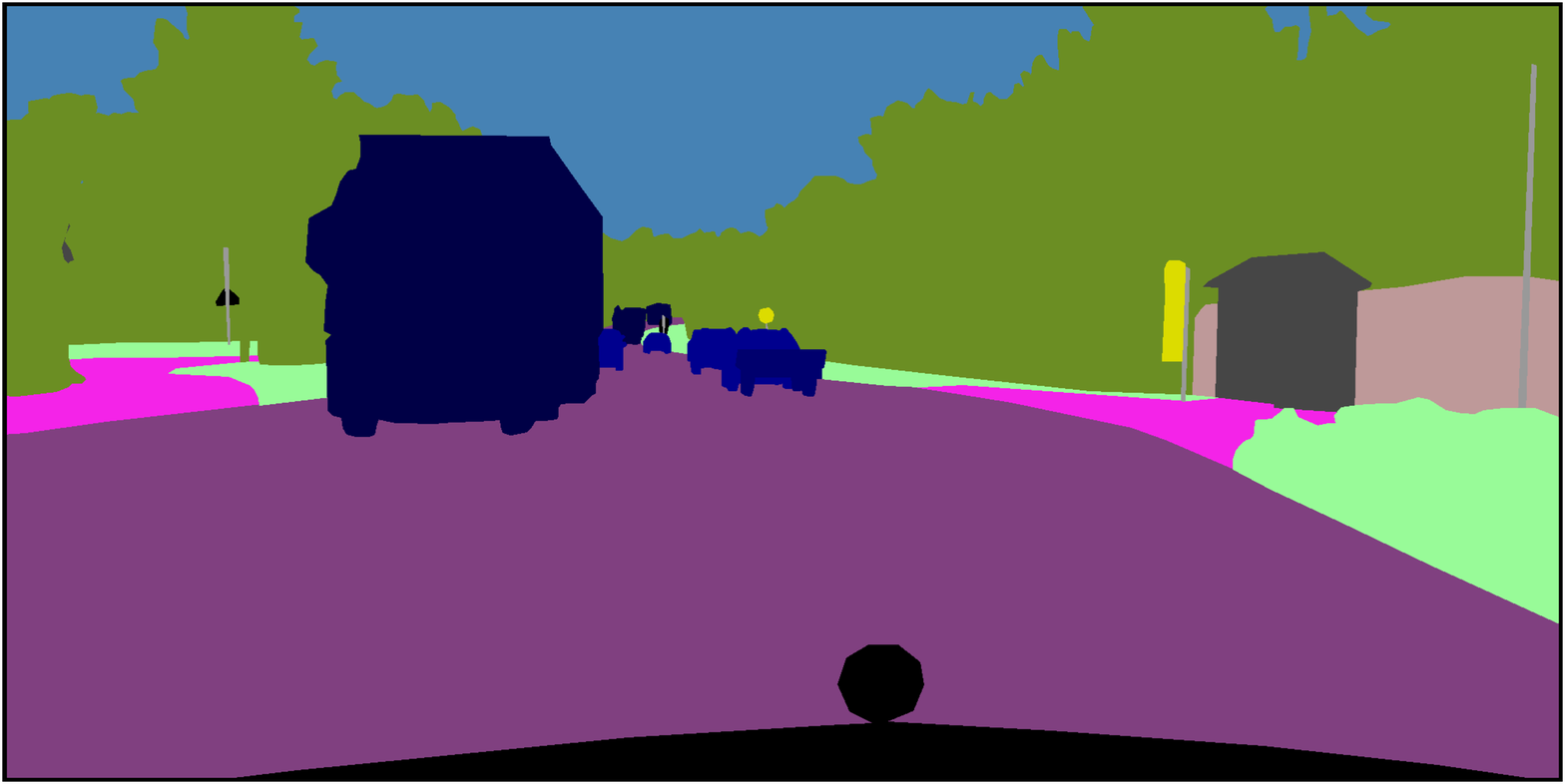}}\hspace{0.01cm}
\subfloat[]{\includegraphics[width=0.22\linewidth]{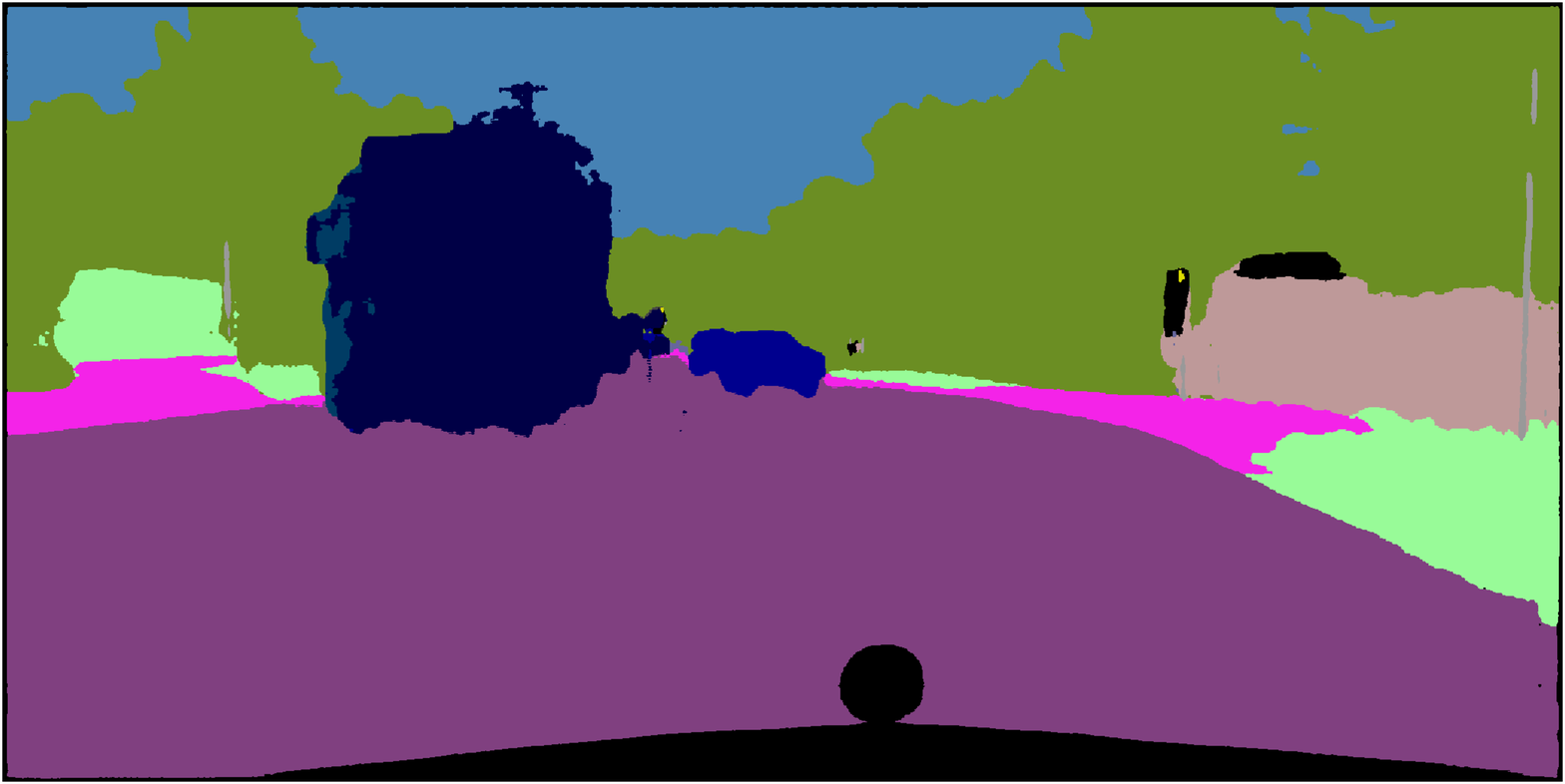}}\hspace{0.01cm}
\subfloat[]{\includegraphics[width=0.22\linewidth]{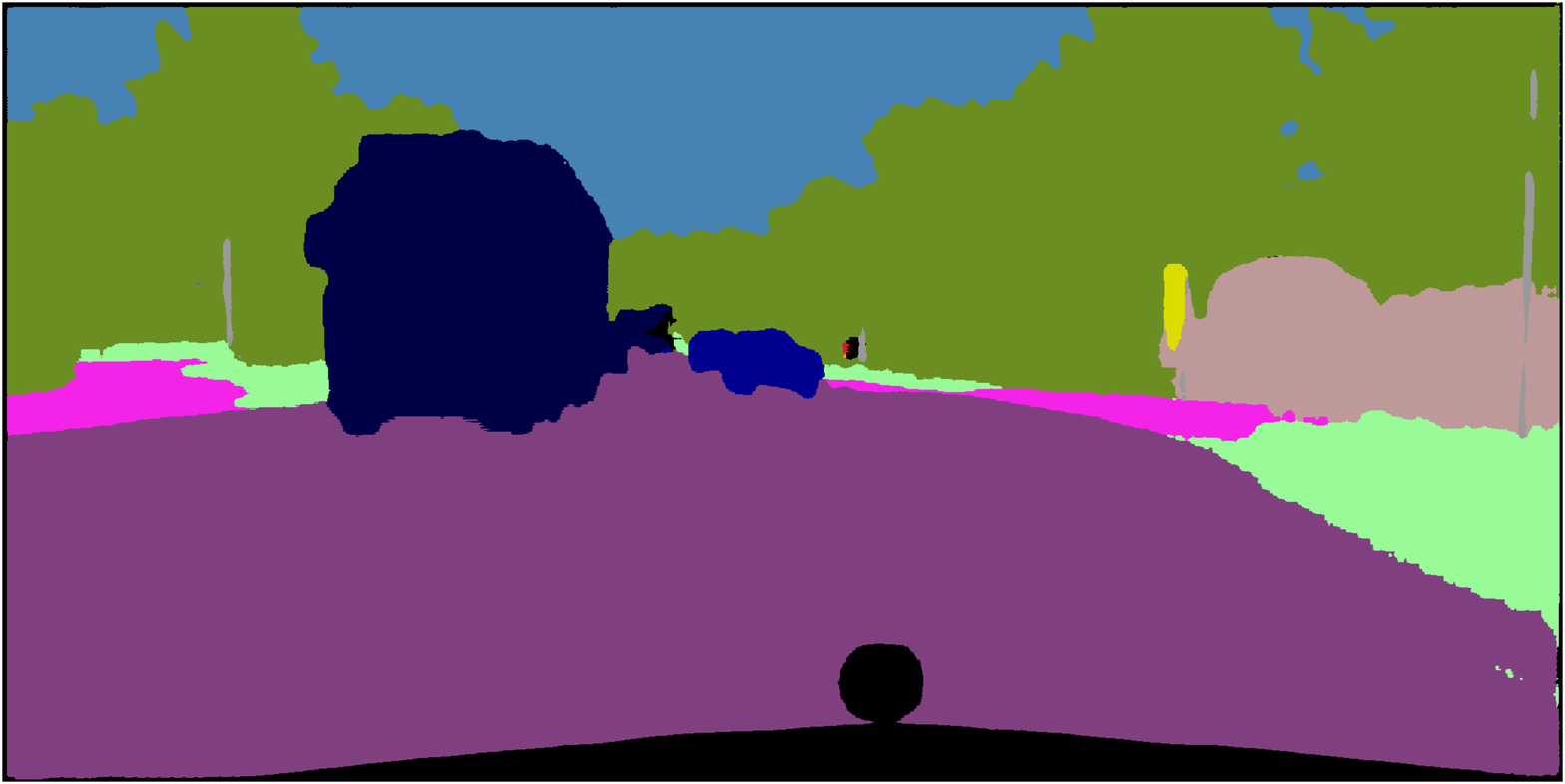}}\hspace{0.01cm}\\
\caption{Qualitative results on Cityscapes validation set: Columns (left-to-right)- input image, grount truth, FLPNet-encoder output and FLPNet output.} \label{output_images}
\end{figure*}

To further scale-down the network size while still maintaining a decent accuracy, we propose an ultra-lightweight real-time FPLNet. Experimental results show that the proposed network achieves 66.93\% and 66.28\% mIoU on validation and test set, respectively. It achieves similar accuracy compared to lightweight models with much less number of parameters such as DABNet \cite{li2019dabnet} and EDANet \cite{lo2019efficient}. When compared to other state-of-the-art ultra-lightweight (less than 0.5 million parameters) networks \cite{mehta2018espnet, paszke2016enet, poudel2018contextnet, wu2020cgnet}, the proposed FPLNet outperforms them in terms of accuracy (almost 1-10\% accuracy improvement) while having similar number of parameters; 0.49 million to be more specific. This shows the effectiveness of our network in achieving a decent accuracy-efficiency tradeoff in Ultra-Lightweight range. More importantly, it achieves accuracy close to large-scale BiSeNetV1 \cite{yu2018bisenet} and mid-scale ERFNet \cite{romera2017erfnet} while being 11.8$\times$ and 4.5$\times$ smaller, respectively. 
To put everything into perspective, a comparison of the proposed FPLNet with other lightweight and ultra-lightweight networks have been shown in Fig. \ref{scatter_plot}. The inherent contention between accuracy and efficiency can also be clearly observed in Fig. \ref{scatter_plot}. For calculating the trainable parameters of the network, pytorch-OpCounter library \cite{orsic2019defense} has been used \cite{nirkin2021hyperseg}. To give a visual perspective, the qualitative results are shown in Fig. \ref{output_images}. \\

\subsubsection{\textbf{Ablation studies}}
In this section, a detailed ablation study of different design choices is presented. It should be noted that all the experiments in this study are conducted considering only the encoder of the FPLNet unless stated otherwise. Once the required efficient encoder was found, a lightweight decoder is attached to it to complete the proposed network, i.e., FPLNet.

\textbf{ESP vs FPL module:}
To show the effectiveness of the proposed FPL module and the corresponding FPLNet, we conducted a series of experiments as shown in Table \ref{module_ablation}. By applying the ESP module in FPLNet and FPL module in ESPNet, we are able to show the effectiveness of both the FPL module and the architecture of our network. It is clear from Table \ref{module_ablation} that by merely replacing ESP module with FPL module in the ESPNet (while keeping other variables constant) we are able to improve the mIoU of ESPNet by 2.43 \%; reflecting the effectiveness of FPL module. Similarly, applying ESP module in our FPLNet (while keeping other variables constant) gives a 2.99 \% accuracy boost over ESPNet; showing the effectiveness of the proposed architecture.

\begin{table}[h]
  \centering
  \small
  \caption{Performance analysis of ESP and FPL module when employed in FPLNet and ESPNet, respectively.}
  \begin{tabular}{l|c|c}
    \hline
     \textbf{Model} & \textbf{mIoU} (\%) & \textbf{Parameters (M)} \tstrut \bstrut \\
     \hline
     ESPNet & 53.30 & 0.34 \tstrut \\ 
     ESPNet-FPL & 55.73 & 0.38 \bstrut \\
     \textbf{FPLNet} & 58.83 & 0.42 \bstrut \\
     FPLNet-ESP & 56.29 & 0.38 \bstrut \\
   \hline  
  \end{tabular}
  \label{module_ablation}
\end{table}

\textbf{Hasty downsampling vs Delayed downsampling:}
In order to reduce parameter contributed by the first stage, many works adopt a hasty downsampling strategy \cite{mehta2018espnet, romera2017erfnet}. This does not allow any feature extraction at stage-1 (i.e., half of input resolution) and perform two back-to-back downsampling operations to downsample the feature maps to 1/4 of input resolution. However, it has an adverse effect on the accuracy of the network as the low-level finer details are lost. Hence, we employ two conventional layers in this stage to extract the finer details which is crucial in fine delineation of the object boundaries in the final segmented map. Table \ref{hasty} shows the advantage of delayed downsampling over hasty one.

\begin{table}[h]
  \centering
  \small
  \caption{Effect of hasty and delayed downsampling on network accuracy and parameters.}
  \begin{tabular}{l|c|c}
    \hline
     \textbf{Downsampling} & \textbf{mIoU} (\%) & \textbf{Parameters (M)} \tstrut \bstrut \\
     \hline
     Delayed & 58.83 & 0.42 \tstrut \\ 
     Hasty & 56.97 & 0.40 \bstrut \\
   \hline  
  \end{tabular}
  \label{hasty}
\end{table}

\textbf{Different fusion strategies within the encoder:}
Table \ref{fir_table} presents a comparison between different variants of intra-stage feature fusion and image fusion. It is clear from the Table \ref{fir_table} that concatenation works better for intra-stage feature fusion as opposed to addition, with a slight increment in the network parameters. The proposed FIR unit

\begin{table}[h]
  \centering
  \small
  \caption{Effect of different types of fusion strategies within the encoder. IF: Image Fusion, ISFF-add: Intra-stage feature fusion by addition, ISFF-concat: ISFF by concatenation.}
  \begin{tabular}{l|cc}
    \hline
     \textbf{Fusion strategies} & \textbf{mIoU} (\%) & \textbf{Parameters (k)} \tstrut \bstrut \\
     \hline
     IF & 57.00 & 414.810 \tstrut  \\
     ISFF-add & 56.42 & 414.636 \bstrut \\
     ISFF-concat & 56.97 & 419.820 \bstrut \\
     IF \& ISFF-add & 57.20 & 414.810 \bstrut \\
     IF \& ISFF-concat & 57.75 & 419.994 \bstrut \\
     \textbf{FIR (proposed)} & 58.83 & 419.994 \bstrut \\
    \hline
  \end{tabular}
  \label{fir_table}
\end{table}

\textbf{Number of FLP modules in different stages:}
A series of experiments have been conducted to find the optimal number of FPL modules to be used in different stages. Table \ref{fplblocks} presents the performance of the FPLNet encoder against different number of FPL modules in stage-2 and stage-3. After performing extensive experiments, it is found that the optimal configuration is 4-8, i.e., 4 and 8 FPL modules in stage-2 and stage-3, respectively.
\begin{table}[h]
  \centering
  \small
  \caption{Effect of varying the number of FPL modules in different stages on the network performance. ``\#FPL\textsubscript{S2}" and ``\#FPL\textsubscript{S3}": Number of FPL modules in stage-2 and stage-3, respectively.}
  \scalebox{0.700}{
  \begin{tabular}{l|cccccccc}
  \hline
  (\#\textbf{FPL\textsubscript{S2}}, \#\textbf{FPL\textsubscript{S3}}) & (2, 6) & (2, 8) & (2, 10) & (2, 12) & (4, 6) &  (4, 8) & (4, 10) & (4, 12) \tstrut \bstrut\\
  \hline
  mIoU (\%)  & 56.0 & 56.1 & 55.9 & 55 & 57.4 & \textbf{58.83} & 57.2 & 57.98  \tstrut \\
  Parameters (M) & 0.32 & 0.40 & 0.47 & 0.57 & 0.34 & \textbf{0.42} & 0.49 & 0.57 \bstrut \\
  \hline
  \end{tabular}
  }
  \label{fplblocks}
\end{table}

\iffalse
\textbf{Channel depth reduction from encoder to decoder (Gradual vs Abrupt):}
In abrupt reduction, the channel depth of encoder output is directly reduced to number of classes (19 in this case), whereas in gradual reduction, channels are reduced gradually (usually by some factor). To reduce the number of parameters, many works reduce the channel depth drastically when switching from encoder to decoder \cite{mehta2018espnet}. This strategy saves some parameters but affects the accuracy significantly as shown in Table \ref{enc2dec}. Our finding is that gradual reduction in channel depth enhances the accuracy with only a slight increase in parameters. To be more specific, the gradual channel reduction employed in FPLNet adds only 0.06 million parameters, while corresponding accuracy gain in 2.81 \%. Hence, we prefer gradual channel reduction in the final network. 

\begin{table}[h]
  \centering
  \small
  \caption{Effect of gradual and abrupt channel depth reduction on the network accuracy. The symbol "$\Rightarrow$" indicates transition from encoder to decoder, whereas "$\rightarrow $" indicates transition from one stage of the decoder to another.}
  \begin{tabular}{l|c|c}
    \hline
     \textbf{Channel reduction} & \textbf{mIoU} (\%) & \textbf{Parameters (M)} \tstrut \bstrut \\
     \hline
     Gradual: 259 $\Rightarrow$ 64 $\rightarrow$ 16 $\rightarrow$ C & 66.93 & 0.49 \tstrut \\ 
     Abrupt: 259 $\Rightarrow$ C $\rightarrow$ C $\rightarrow$ C & 64.12 & 0.43 \bstrut \\
   \hline  
  \end{tabular}
  \label{enc2dec}
\end{table}
\fi

\textbf{Extent of factorization:}
Factorization of convolutional kernals saves parameters but affects the accuracy. Hence, a trade-off is required. To find the required trade-off, a set of experiments have been conducted to reveal how much factorization within the FPL module should be done. Table \ref{factorization} presents the results of the corresponding experiments. 
\begin{table}[h]
  \centering
  \small
  \caption{Effect of factorization of conv kernals in FPL module. \ding{55} means no factorization, "All" means all the kernals are factorized including the top convolution of pyramid stage-1.}
  \begin{tabular}{c|cc}
    \hline
     \textbf{Factorization} & \textbf{mIoU} (\%) & \textbf{Parameters (M)} \tstrut \bstrut \\
     \hline
     \ding{55} & 59.10 & 0.54 \tstrut \\
     All & 56.79 & 0.38 \bstrut \\
     \textbf{Proposed (composite)} & 58.83 & 0.42 \bstrut \\
    \hline
  \end{tabular}
  \label{factorization}
\end{table}
In the proposed composite scheme, first stage of pyramid, i.e., $3 \times 3$ convolution is kept symmetric while the second stage, i.e., bank of dilated convolutions are factorized. It is clear from Table \ref{factorization} that the proposed setting of factorization in the FPL module results in a balanced trade-off. With no factorization at all, we observe only 0.38 \% mIoU increment at the expense of 0.13 million more parameters compared to the composite scheme. Hence, we can easily conclude that the proposed configuration of factorization offers the optimal balance between mIoU and number of parameters.

\textbf{Activation function:}
Using PReLU \cite{he2015delving} instead of ReLU \cite{nair2010rectified} results in 0.45 \% increment in accuracy, while having a negligible effect on the model complexity. So, we adopted PReLU in our network.

\section{Conclusion and Future work} \label{conclusion}
In this article, a novel module has been proposed which employs spatial pyramid to extract multi-scale context in an efficient way. The resulting module is called factorized pyramidal learning (FPL) module. To allow more information embedding along the channel dimension, the spatial pyramid is decomposed into two stages. The first stage is a conventional convolution where as the second stage employs bank of factorized convolutional filters with different dilation rates. As a result, it is able to capture both short-range and long-range context which greatly enhances the segmentation accuracy of the model. The proposed FPL module carefully factorizes the pyramid filters, resulting in a huge saving of overall trainable parameters. Moreover, to improve the information flow and to enable enhanced learning, shallow and deep features of each stage is fused with the downsampled image using a dedicated Feature-Image Reinforcement (FIR) unit. This gives accuracy boost without increasing the network parameters compared to simple concatenative fusion. To complete the network, we also design a small, simple and sequential asymmetric decoder for the recovery of local spatial details in the final segmentation map. Based on the FPL module, FIR unit and the asymmetric decoder, we propose a lightweight real-time network to achieve state-of-the-art accuracy-efficiency trade-off. A detailed ablation study is presented to provide deep insight of the network response against various design choices. 

\bibliographystyle{IEEEtran}
\bibliography{reference.bib}

\end{document}